%% file: main.tex
\documentclass[journal]{IEEEtran} 
\usepackage{graphicx} 
\input{preamble}

\begin{document}

\title{%
Robot Learning on Discrete Surfaces: \\ Theory and Applications
\thanks{Co-funded  by the European Union projects INVERSE (grant agreement no. 101136067) and MAGICIAN (grant agreement no. 101120731)}}

\author{Matteo Dalle Vedove$^{1,2}$, Fares J. Abu-Dakka$^{3}$, Luigi Palopoli$^{4}$, Daniele Fontanelli$^{1}$, Matteo Saveriano$^{1}$
\thanks{$^{1}$Department of Industrial Engineering, Universit\`a di Trento, Trento, Italy. \tt\small matteo.dallevedove@unitn.it}
\thanks{$^{2}$DRIM, Ph.D. of national interest in Robotics and Intelligent Machines.}
\thanks{$^{3}$Mechanical Engineering Program, Division of Engineering, New York University Abu Dhabi, Abu Dhabi, United Arab Emirates.}
\thanks{$^{4}$Department of Information Engineering and Computer Science, Universit\`a di Trento, Trento, Italy.}
}

\markboth{IEEE Transactions on Robotics,~Vol.~XX, No.~XX, February~2026}%
{Dalle Vedove \MakeLowercase{\textit{et al.}}: xxx}

\maketitle

\begin{abstract}
All the objects composing our world are enclosed within surfaces. Yet, most robot learning and motion generation frameworks treat surfaces as constraints ignoring their intrinsic geometry. This gap is acute for polyhedral meshes---the standard output of CAD and 3D reconstruction---whose discrete geometric structure remains unexploited.
In this paper, we propose a unified discrete Riemannian framework that enables robot learning directly on polyhedral surface meshes. Using discrete differential geometry, we define logarithmic and exponential maps, parallel transport, and ambient-space projections that remain well-defined across faces, edges, and vertices.
We instantiate the framework in three learning paradigms: (\emph{i}) Dynamic Movement Primitives (DMPs), an improved exponential-map computation and a fixed-tangent-cone forcing-term encoding with parallel transport yield better cross-surface generalisation and stability over prior mesh-based approaches. (\emph{ii}) Gaussian Process (GP), a geodesic-based kernel with practical admissibility control, enables regression at arbitrary mesh locations without smoothness assumptions. (\emph{iii}) Riemannian Flow Matching (RFM), mesh-native operators improve generative quality over spectral baselines while reducing training time.
The framework is validated in simulation against state-of-the-art methods and demonstrated on two real-robot scenarios: generalising user-drawn trajectories across different surfaces and planning polishing motions on RGB-D-reconstructed surfaces.
\end{abstract}


\maketitle

\input{Sections/introduction.tex}
\input{Sections/related-works.tex}
\input{Sections/background.tex}
\input{Sections/methodology.tex}

\input{Sections/simulation.tex}
\input{Sections/experiments.tex}
\input{Sections/conclusions.tex}

\bibliographystyle{ieeetr}
\bibliography{references}

\end{document}

%% file: preamble.tex
\usepackage[dvipsnames]{xcolor}
\usepackage{amsfonts}
\usepackage{amsmath}
\usepackage{amssymb}
\usepackage{tikz, tikz-3dplot} 
\usepackage{subfigure} 
\usepackage{hyperref} 
\usepackage{booktabs} 
\usepackage{enumerate} 
\usepackage{upgreek}
\usepackage{algorithm}
\usepackage{algpseudocode}
\usepackage{cite}
\usepackage{standalone}

\usetikzlibrary{calc,intersections,arrows,arrows.meta,angles,quotes,decorations.pathmorphing,decorations.pathreplacing}

\newcommand{\interior}[1]{%
  {\kern0pt#1}^{\mathrm{o}}%
}

\renewcommand{\epsilon}{\varepsilon}

\newcommand{\R}{\mathbb R}

\newcommand{\M}{\mathbb M}

\newcommand{\Exp}{\mathrm{Exp}}
\newcommand{\Log}{\mathrm{Log}}
\newcommand{\GP}{\mathcal{GP}}
\newcommand{\tri}{\mathcal{T}}
\newcommand{\face}{\tri}

\newcommand{\mesh}{\mathcal{M}}
\newcommand{\vertexset}{\mathcal{V}}
\newcommand{\faceset}{\mathcal{F}}
\newcommand{\g}{\mathfrak{g}}
\renewcommand{\d}{\mathrm{d}}

\newcommand{\expected}[1]{\mathbb{E}\left\{#1\right\}}
\newcommand{\covariance}[1]{\mathbb{C}\textrm{ov}\left\{#1\right\}}

%% file: Sections/introduction.tex
\section{Introduction}
\IEEEPARstart{I}n real applications, robots  have to learn how to physically interact with objects and their environment. Sometimes this is required for manipulation purposes (e.g., grasping, pick-and-place, insertion), and sometimes for manufacturing operations (e.g., wiping, finishing, drawing, grinding, cutting, inspecting). A common trait of all these operations is that they are not executed in free space: in the vast majority of cases, they entail continuous contact with a surface enclosing the target object, where friction effects and geometric irregularities play a significant role. This raises a fundamental question: can a learning process for such tasks be truly effective if the geometric properties of the contact surface are ignored?
This issue has been often overlooked in the literature. Most existing Learning from Demonstration (LfD) and motion generation frameworks assume that skills evolve in flat Euclidean spaces and can be directly transferred to surfaces. When task complexity increases, surface constraints are typically recovered a posteriori through projection methods~\cite{Sentis2005}, impedance control, or local parameterisation. While effective in structured and repeatable settings, treating surface geometry as an external constraint---rather than an intrinsic component of the learning process---often results in poor generalisation when the underlying structure changes.

In recent years, geometry-aware learning methods have been proposed for some classes of Riemannian manifolds, such as rotations or symmetric positive-definite matrices~\cite{Calinon20RAM,Jaquier2021,Chen2023,AbuDakka20}, for which researchers have developed tools and methods to learn robot skills in non-Euclidean spaces. These formulations rely on differential operators---such as exponential and logarithmic maps, and parallel transport---that enable learning and control directly in the tangent space. However, extending these methods to surface-based skills remains an open issue, as surfaces encountered in real applications are rarely available as smooth manifolds. In most cases, surfaces are obtained from CAD models, perception pipelines, or 3D reconstruction tools~\cite{Berger2017}, for which the most natural representation is a 3D polyhedral mesh. Such discrete representations exhibit non-differentiable singularities at edges and vertices and do not admit the smooth structure assumed by classical Riemannian learning frameworks.

The absence of a smooth representation creates a significant gap between geometry-aware robot learning theory and its practical applicability to discrete manifolds. Although surfaces can be studied through the lens of differential geometry, constructing smooth representations for objects with arbitrary topology is generally intractable~\cite{Hubeli2000}. At the same time, naïve discretisation of smooth manifold operators on meshes either requires restrictive assumptions on surface smoothness or leads to inconsistencies and ill-defined behaviour at singularities. As a result, existing approaches often rely on local approximations, ad hoc embeddings, or spectral smoothness assumptions that limit robustness and hinder generalisation~\cite{Pantic2021, DalleVedove25, Borovitskiy2020}.

\noindent 
\textbf{Paper Contribution. } In this work, we take a more direct route to fill this gap by developing a discrete Riemannian framework for robot learning directly on polyhedral surface meshes. Starting from discrete differential geometry~\cite{Bobenko2023}, we construct
geometrically consistent approximations of key Riemannian
operators---including logarithmic and exponential maps, parallel
transport, geodesic distance, and ambient-space projections---that remain well-defined across faces, edges, and vertices of a mesh. Unlike prior mesh-based formulations~\cite{DalleVedove25}, which leave edge and vertex singularities unaddressed and rely on local  approximations of the exponential map (valid only for small displacements), our operators are built on the straightest-geodesic principle~\cite{Polthier2006} and are equipped with tangent-cone projections that handle all singular loci. Building on this geometric foundation, we show how three established learning frameworks can be naturally extended to discrete surfaces: (\emph{i}) geometry-aware Dynamic Movement Primitives (DMPs) with a fixed-tangent-cone forcing-term encoding that replaces the local parameterisations used in~\cite{DalleVedove25, Xue2023}, (\emph{ii}) geodesic-distance-based
Gaussian Processes (GPs) with controllable kernel admissibility, and (\emph{iii}) Riemannian Flow Matching (RFM) with mesh-native operators replacing the spectral premetrics of~\cite{Chen2023}. These formulations yield motion and function representations that generalise across surfaces with different curvature, topology, and discretisation resolution, removing the need for smoothness or spectral assumptions.

To summarize, our main contributions are as follows:
\begin{itemize}
\item A discrete Riemannian operator framework on polyhedral meshes---including tangent cones, straightest-geodesic exponential maps, and ambient-space projections---that explicitly handles edge and vertex singularities left unaddressed by existing mesh-based
  methods~\cite{DalleVedove25,Pantic2021}, enabling consistent geometric computations directly on discrete surface representations like polyhedral meshes.
\item A reformulation of geometry-aware DMPs on meshes that encodes forcing term in a fixed tangent cone and reconstructs it via parallel transport during execution, replacing the local, velocity-aligned parameterisation of~\cite{DalleVedove25} and eliminating the limit-cycle drift it induces on curved surfaces.
\item A geodesic-distance-based GP formulation on meshes that enables kernel regression at arbitrary surface locations---including face interiors---without smoothness or density requirements, overcoming the vertex-restriction and resolution-sensitivity of Laplace–Beltrami-based kernels~\cite{Borovitskiy2020}.
\item An extension of RFM to discrete polyhedral surfaces using geodesic premetrics, improving generative quality and reducing training time relative to the spectral baselines of~\cite{Chen2023}.
\end{itemize}

The rest of the paper is organised as follows. Section~\ref{sec:related} analyses the state of the art. After a quick recall of foundation knowledge (Sec.~\ref{sec:background}), we present the unified approach for geometry-aware learning on discrete surfaces in Sec.~\ref{sec:method}. Section~\ref{sec:simulations} provides simulation results and comparisons with the state-of-the-art, while Sec.~\ref{sec:experiments} provides experiments on a real robot highlighting the strengths of our approach. We conclude in Sec.~\ref{sec:conclusion} with some limitations of the approach and final remarks.

%% file: Sections/related-works.tex
\section{Related work}
\label{sec:related}

\paragraph*{Geometric Learning}
Learning from Demonstration (LfD) is a widely adopted paradigm for encoding human expertise and transferring it into robotic systems~\cite{billard2016learning}. Classical LfD approaches---including DMPs~\cite{saveriano2023dmp}, Probabilistic MPs~\cite{paraschos2013}, Kernelised MPs~\cite{huang2019kernelized}, or Gaussian Mixture Models (GMMs)~\cite{calinon2016tutorial}---have been successfully applied to learn skills defined in Euclidean spaces. However, these methods do not account for the intrinsic geometry of constrained domains, and consequently fail when applied to data evolving on non-linear spaces such as rotations, stiffness matrices, or surfaces. To address this limitation, Lie-group-based formulations and Riemannian geometry have been introduced, with Lie-algebra approaches proving effective for specific spaces~\cite{Ude14, AbuDakka2020-vicl, Koutras20a, Kim2025}, and Riemannian formulations offering greater flexibility by imposing fewer structural restrictions on the domain~\cite{AbuDakka24, Jaquier2021, Calinon20RAM, Chen2023, Ratliff2018}. As a result, geometry-aware learning methods on smooth manifolds have been successfully applied to rotations, symmetric positive-definite matrices, and Grassmannian manifolds, where closed-form expressions of geometric operators are available. However, extending these results to surfaces remains largely an open problem, as real-world surface representations lack the fundamental structure that these methods rely on.

\paragraph*{Learning on smooth and discrete manifolds}
Several methods have been proposed to parameterise smooth surfaces using spline or subdivision representations~\cite{Dimas1999-cp, Boier2004, Choi2020}. While effective in computer graphics and geometric modelling, these approaches typically provide only local parameterisations or impose strong constraints on surface topology, restricting their applicability to robotic learning tasks involving complex or reconstructed environments. Working directly on polyhedral surface meshes has emerged as a promising alternative. Pantic et al.~\cite{Pantic2021} proposed a mesh-manifold-based motion planning algorithm, framed as a Riemannian motion policy~\cite{Ratliff2018}, that uses isomorphic mappings to flatten meshes onto planar domains. However, this strategy requires restrictive topological assumptions and implicitly assumes sufficient mesh smoothness, avoiding the singularities arising at edges and vertices rather than addressing them. 

The MeshDMP framework~\cite{DalleVedove25} represents the most direct prior reference for our approach, as it extends geometry-aware DMPs~\cite{AbuDakka24} to discrete surfaces. While MeshDMP removes topological restrictions on the mesh, it suffers from two important limitations. First, it relies on an approximate exponential map that is valid only under bounded face-angle conditions and does not explicitly address vertex singularities. Second, its forcing-term encoding is defined in a local, moving coordinate system, which can lead to instability and degraded generalisation, particularly for rhythmic motion patterns.

Our approach addresses both limitations of MeshDMP and, more broadly, establishes a geometrically consistent foundation for learning on discrete surfaces. By adopting a straightest-geodesic-based~\cite{Polthier2006} exponential map and explicitly handling edge and vertex singularities, we obtain a more principled and consistent geometric formulation. Additionally, by encoding the forcing term in a fixed tangent space and transporting it during execution, we achieve substantially improved stability and transferability across surfaces, without relying on local parameterisations.

\paragraph*{Meshes and the Laplace-Beltrami operator}
The Laplace-Beltrami operator (LBO) is the generalisation of the Laplacian to smooth Riemannian manifolds. Despite being initially developed for differentiable structures, its definition has been extended to discrete domains, including polyhedral meshes~\cite{Vallet2008-al, Crane2013-hp}. The LBO is a powerful tool for learning on surfaces, as its spectral decomposition can be used to approximate distances, kernels, and function spaces~\cite{Patane2014, Koestler2022, Lang2023}. As an example, a pioneering work by Borovitskiy et al.~\cite{Borovitskiy2020} exploited the eigenfunction basis of the LBO to construct Mat\'ern kernels on compact manifolds, effectively enabling Gaussian Process Regression (GPR) on meshes. 

A key advantage of LBO-based methods is that they offer theoretical guarantees. However, they implicitly assume that the mesh is a sufficiently good approximation of a smooth surface and restrict function evaluation to mesh vertices. As a consequence, accurate GPR often requires dense meshes or additional interpolation steps, increasing computational cost and approximation error. Our approach, instead, relies directly on geodesic distance computation, enabling GPR at arbitrary surface locations---including points in the interior of faces---without any smoothness requirement. It is worth mentioning that geodesic-based kernels may lose positive definiteness beyond a critical lengthscale~\cite{Feragen2015}. We address this limitation by treating it as a practical modelling trade-off and introduce criteria that ensure stable GPR in realistic scenarios.

\paragraph*{Riemannian generative models}
Continuous Normalising Flows (CNFs)~\cite{Chen2018} are generative models that represent probability distributions by learning smooth, time-dependent vector fields whose integration defines a continuous transformation from a simple base distribution to a target data distribution. Despite their theoretical appeal and exact likelihood formulation, CNF-based architectures are computationally expensive to train, particularly in high-dimensional settings, and can suffer from numerical instability. Conditional Flow Matching (CFM)~\cite{Lipman2023, Liu2023} addresses these limitations by replacing likelihood maximisation with a regression-based objective, substantially improving both training stability and efficiency. 

Particularly relevant for geometric learning is the work by Chen et al.~\cite{Chen2023}, who extend CFM to Riemannian manifolds (RFM) by explicitly incorporating the geometric structure into the flow definition. By defining probability paths on manifolds, their approach trains a neural network to learn vector fields using exact differential operators on analytically defined manifolds (e.g., hyperspheres and tori), or approximate operators for discrete domains such as meshes, constructed via premetrics induced by the Laplace–Beltrami operator. 

Our framework extends RFM to polyhedral meshes by introducing geometrically consistent operators tailored to mesh representations, enabling more stable and accurate vector field learning. As demonstrated in the evaluation below, the performance improvement is consistent across the tested baseline methods.

%% file: Sections/background.tex
\section{Background}
\label{sec:background}
In this section, we present some background notions needed for a proper understanding of our approach. The notation used throughout the paper is reported in Table~\ref{tab:notation}.

\input{Tables/symbols.tex}

\subsection{Riemannian Manifolds} \label{sec:riemann-manifolds}

A Riemannian $n$-manifold is a pair $(\M, \g)$, where $\M$ is a
differentiable manifold and
$\g: T_{(\cdot)} \M \times T_{(\cdot)} \M \rightarrow \R$ is a
smoothly varying inner product defined on each tangent space
$T_{(\cdot)}\M$~\cite{doCarmoRiemann}.  Based on the definition of the
metric $\g$, one can construct other meaningful concepts, such as
distances, angles, volumes, and curvature.

The choice of $\g$ plays a critical role and it is application dependent. In this work, we consider the metric to be the one induced by the ambient space of the manifold, i.e.,
\begin{equation}
    \g: \langle v, w\rangle \triangleq v^\top w \quad v, w \in T_y\M \subset \R^{2n},
\end{equation}
where we define the inner product using the dot product in Euclidean (ambient) space. 
This is always possible since any smooth 
$n$-manifold admits a smooth embedding into $\R^{2n}$~\cite{Hassler1944},  which implies that, for any point $p \in \M$, the tangent plane $T_p\M$ is a $n$-dimensional linear subspace of $\R^{2n}$.

Differential geometry is built around the notion that a manifold $\M$
locally resembles its tangent space $T_{(\cdot)}\M$, which provides a
Euclidean approximation at each point.  For this reason, specific
operators are required to map values from manifolds to tangent spaces,
and vice versa.  Let $p \in \M$ be a point of interest on the
manifold.  The logarithmic map $\Log_p: \M \rightarrow T_p\M$ is the
differential operator which enables the expression of any other point
on $\M$ in the tangent space $T_p\M$ of $p$.  The dual operation of
mapping values from the tangent space back to the manifold is achieved
through the exponential map $\Exp_p: T_p\M\rightarrow \M$.  Finally,
the parallel transport $P_{p\rightarrow q}: T_p\M \rightarrow T_q\M$
is the operator that moves vectors from the tangent space of $p$ to
the tangent space of another point $q\in \M$, while preserving their
geometric properties (such as length and angle) along the curve
connecting $p$ and $q$ on $\M$.
We refer to~\cite{Calinon20RAM} for the definition, as well as for a
good visual and intuitive representation, of these operators for
popular manifolds in robotics.

\subsection{Polyhedral Meshes}

A polyhedral mesh is a discrete representation of a surface formed by a finite 
collection of planar polygonal faces, typically triangles, joined along shared 
edges and vertices.
This representation is widely used in Computer-Aided Design (CAD) and computer 
vision, and is supported by a wide range of data structures and algorithms 
available off-the-shelf in the form of libraries~\cite{DeFloriani2005}.

In this work, we focus on manifold surface meshes $\mesh$, which can be viewed 
as piecewise-linear, two-dimensional manifold surfaces embedded in $\R^3$.
In contrast to the smooth surfaces of differential geometry, polyhedral meshes 
are flat on each face and exhibit discontinuities in the normal field at edges 
and vertices, where tangent planes are not uniquely defined.
For this reason, polyhedral meshes are commonly referred to as \textit{discrete 
surfaces}.

\begin{figure}[bt]
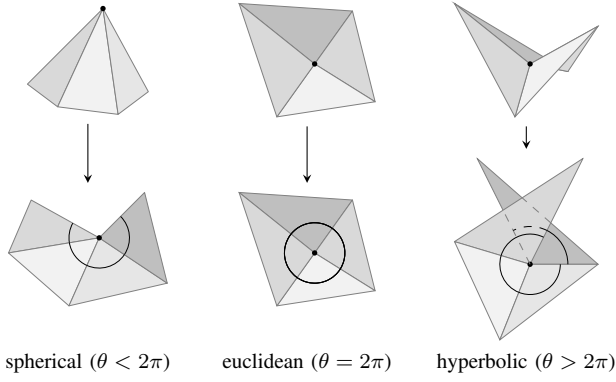

    \centering
    \includestandalone{Figures/angles}
    \caption{
    Examples of spherical, euclidean, and hyperbolic vertices of a polyhedral mesh.
    }
    \label{fig:vertexclass}
\end{figure}

A mesh $\mesh$ is described as a graph whose nodes are the vertices
$\nu_i \in \vertexset$, connected by edges, and whose faces
$\face_i \in \faceset$ are the planar polygons bounded by those edges;
without loss of generality, we consider faces to be triangles.
Relevant to our work is the definition of the total angle $\theta(p)$
for points $p\in \mesh$.  Following~\cite{Polthier2006}, the total
angle for points which are either in the interior or on the edge of a
triangular face is $\theta(p) \triangleq 2\pi$, since the surface is
locally isometric to the Euclidean space at those points; at a vertex
$\nu_i$, we define
\begin{equation} \label{eq:totalangle}
    \theta(\nu_i) \triangleq \sum_{\face_j \in \mathcal{I}(\nu_i)} 
    \theta_{\face_j}(\nu_i),
\end{equation}
where $\mathcal{I}(\nu_i)$ is the set of all faces adjacent to vertex
$\nu_i$, and $\theta_{\face_j}(\nu_i)$ is the angle subtended by face
$\face_j$ at vertex $\nu_i$.  Based on the total angle, points are
classified as spherical ($\theta(\nu_i) < 2\pi$), Euclidean
($\theta(\nu_i) = 2\pi$), or hyperbolic ($\theta(\nu_i) > 2\pi$), as
shown in Fig.~\ref{fig:vertexclass}.

\subsection{Geometry-Aware Dynamic Movement Primitives}
Dynamic Movement Primitives (DMPs) are an established framework to learn a 
generalisable representation of sensorimotor skills from demonstrations, by 
encoding expert demonstrations into a dynamical system~\cite{Ijspeert13, 
saveriano2023dmp}.
Initially developed to operate in Euclidean space, DMPs have sparked interest 
throughout the robotics community, which has proposed their extension to 
different non-Euclidean manifolds.

Among different available formulations, we recall the discrete geometry-aware DMP~\cite{AbuDakka24} which uses the Riemannian operators on smooth manifolds:
\begin{equation} \label{eq:gdmp}
    \begin{aligned}
        \tau \nabla_z z & = \alpha ( \beta \Log_p(g) - z) + f(x),  \\
        \tau \dot p & = z, \\
        \tau \dot x & = -\zeta x.
    \end{aligned}
\end{equation}
Here, is the position state (point on the manifold), $g \in M$ is the dynamics attractor, $z \in T_pM$  is the scaled velocity state, $x \in \R$ is a time encoding variable, and $f: \R \rightarrow T_pM$ is the forcing function which is learned from a demonstration, and is often modelled
as a radial basis functions network. 

\subsection{Gaussian Processes on Manifolds}
GPs are powerful and flexible probabilistic models that define, in a principled 
manner, distributions over functions conditioned on observed data~\cite{Rasmussen2005}.
Let $h: M \rightarrow \R$ be a real-valued random function defined over a 
manifold $M$\footnote{Generally speaking, $h$ may be defined on an arbitrary 
input space $\mathcal{X}$; in this work we are interested in the cases where 
$\mathcal{X}$ is a manifold.}.
The function $h$ is said to follow a GP with mean function $\mu$ and covariance 
function $k$, i.e.,
\begin{equation} \label{eq:gp}
    h \sim \mathcal{GP}(\mu, k),
\end{equation}
if, for any finite set of $m$ input points
$P = \{p_i \in M\}_{i=1}^m$, the corresponding function values
$H = h(P) \in \R^m$ are jointly Gaussian distributed with mean
$\mu_{P} = \mu(P) \in \R^m$ and covariance matrix
$K_{PP} = k(P, P) \in \R^{m\times m}$.  The key component of a GP is
the kernel function $k: M \times M \rightarrow \R$, which encodes
prior information and assumptions about the smoothness and structure
of the underlying function.

Gaussian process regression (GPR) is the technique that exploits GPs to perform 
principled probabilistic predictions of the function value outside the training 
dataset.
Given some observed input-output pairs $(p_i, y_i)$, with $y_i = h(p_i) + 
\epsilon_i$ and $\epsilon_i \sim \mathcal{N}(0, \sigma^2)$, and a prior 
$h \sim \mathcal{GP}(0, k)$, the posterior distribution $h \mid Y$ is another 
GP with the following mean and covariance:
\begin{align}
    \expected{h \mid Y} & = K_{(\cdot)P} \big(K_{PP} + \sigma^2 I\big)^{-1} Y, 
    \label{eq:gp-prediction} \\
    \covariance{h \mid Y} & = K_{(\cdot, \cdot)} - K_{(\cdot)P} 
    \big(K_{PP} + \sigma^2 I\big)^{-1} K_{P(\cdot)},
\end{align}
with $Y = (y_i)$ and $(\cdot)$ denoting arbitrary test locations.

\subsection{Riemannian Flow Matching}
\label{sec:back-rfm}
RFM~\cite{Chen2023} is a generative modelling framework that extends
Continuous Normalising Flows (CNFs) to data supported on Riemannian
manifolds in a geometrically consistent manner.  CNFs learn a smooth,
invertible transformation that maps a simple reference distribution
into a complex target distribution; RFM generalises this idea to
curved spaces by ensuring that the transformation respects the
geometry of the manifold.  Let $\pi: M \rightarrow \R^+$ be a
probability density function on a manifold, and let $\mathcal{P}$ be
the space of probability densities.  RFM seeks to find a probability
path $\pi_t: [0,1] \rightarrow \mathcal{P}$ that continuously deforms
a simple reference distribution $\pi_\textrm{ref} = \pi_0$ into a
target data distribution $\pi_\textrm{data} = \pi_1$.  Such a path is
induced by the flow $\psi_t: M \rightarrow M$ of a time-dependent
vector field $u_t: \R \times M \rightarrow T_{(\cdot)}M$, defined as the
solution of

\begin{equation} \label{eq:flow-ode}
    \frac{\d}{\d t} \psi_t(p) = u_t \big( \psi_t(p)\big), \quad \psi_0(p) = p.
\end{equation}
Intuitively, $\psi_t$ moves particles on the manifold along trajectories 
dictated by the vector field $u_t$, progressively transporting mass from 
$\pi_\textrm{ref}$ to $\pi_\textrm{data}$.
The RFM objective is to learn a parametric vector field $v_t(p, \Theta) \in 
T\M$ ---generally using Neural Networks (NNs)--- which best approximates $u_t$ 
by minimising
\begin{equation} \label{eq:rcfm}
    \mathcal{L}_\textrm{RFM}(\Theta) = \mathbb{E}_{t \sim \mathcal{U}(0, 1), 
    p \sim \pi_t} \left\| v_t(p, \Theta) - u_t(p) \right\|^2_{\mathfrak{g}},
\end{equation}
where the norm $\|\cdot\|_{\mathfrak{g}}$ is induced by the Riemannian metric, 
ensuring that the discrepancy between vector fields is measured in a 
geometrically meaningful way.
Once $v_t(\cdot, \cdot)$ is learned, samples from $\pi_\textrm{data}$ are 
generated by integrating the flow starting from $p \sim \pi_\textrm{ref}$.
The main challenge lies in the construction of the supervision signal $u_t$.
In FM, a premetric $D: M \times M \rightarrow \R$ is a bivariate function 
on the manifold which is positive definite ($D(p, q) \geq 0\ \forall\, p,q 
\in M$ and $D(p, q) = 0 \Leftrightarrow p = q$) and non-degenerate 
($\nabla_p D(p, q) \neq 0 \Leftrightarrow p \neq q$); intuitively, $D$ 
generalises the notion of distance without requiring the full structure of 
a metric.
Given a monotonically decreasing scheduler $\eta: [0, 1] \rightarrow \R$ 
with $\eta(0) = 1$ and $\eta(1) = 0$, the FM objective defines a flow 
$\psi_t$ that shrinks the premetric $D(\cdot, p_1)$ according to the law
\begin{equation} \label{eq:rfm-premetric-shrink}
    D\big( \psi_t(p_0 | p_1), p_1 \big) = \eta(t) D(p_0, p_1),
\end{equation}
meaning that particles are steered monotonically closer to their target 
$p_1$ as $t$ progresses.
Chen et al.~\cite{Chen2023} show that such a flow can be obtained by 
defining the conditional vector field $u_t$ as
\begin{equation} \label{eq:fm-vectorfield}
    u_t(p | p_1) = \frac{\d \log \eta(t)}{\d t}\, D(p, p_1) 
    \frac{\nabla D(p, p_1)}{\| \nabla D(p, p_1) \|^2_{\mathfrak{g}}}.
\end{equation}

%% file: Tables/symbols.tex
\begin{table}
    \centering
    \caption{Notation}
    \newcommand{\mysub}{{(\cdot)}}
    \label{tab:notation}
    \begin{tabular}[t]{l l}
        \toprule
        $\M$ & smooth manifold \\
        $\mesh$ & polyhedral mesh \\
        $M$ & manifold (discrete or smooth) \\
        $\g$ & metric on $M$ \\
        $T_{\mysub}\M$ & tangent space on $\M$ at $\mysub$ \\
        $T_{\mysub}\mesh$ & tangent cone on $\mesh$ at $\mysub$ \\
        $TM$ & tangent bundle, set $\left\{ (p, v) \ | \  p \in M, v \in T_p M \right\}$ \\
        $\GP$ & Gaussian Process \\
        $\gamma \subset M$ & geodesic curve \\
        $\ell \in \mathbb{R}^+$ & geodesic length \\
        $\Log_{\mysub}$ & logarithmic map;  $M \mapsto T_{\mysub}M$ \\
        $\Exp_{\mysub}$ & exponential map;  $T_{\mysub}M \mapsto M$ \\
        $P_{\mysub \rightarrow (\circ)}$ & parallel transport;  $T_{\mysub}M \mapsto T_{(\circ)}M$ \\
        $\nabla_{\mysub}$ & covariant derivative; $T_{\mysub}M \times T_{\mysub}M \mapsto T_{\mysub}M$ \\
        $p, q$ & points in $M$ \\
        $v, w, z$ & tangent vectors in $T_{\mysub}M$ \\
        $n$ & normal vector field on $\mesh$ \\
        $g$ & DMP goal/centre in $\mesh$ \\
        $f$ & DMP forcing term \\
        $x$ & DMP phase variable \\
        $\tau, \Omega, \alpha, \beta, \zeta$ & DMP coefficients in $\mathbb{R}^+$ \\
        $h$ & random function drawn from a GP \\
        $k$ & kernel function; mapping $M \times M \rightarrow \R^+$ \\
        $P = \{p_i\}_{i=1}^m$ & collection of $m$ points on $M$ \\
        $y$ & observed sample from $h$ with noise $\mathcal N(0, \sigma^2)$ \\
        $K_{XY}$ & covariance function $k(X, Y)$ \\
        $\kappa$ & kernel length scale \\
        $\pi$ & probability density function on $M$ \\
        $\pi_\textrm{ref}, \pi_\textrm{data}$ & reference, data probability density functions \\
        $\mathcal P$ & space of probability densities \\
        $\phi_t$ & flow of a vecto field $u_t$ \\
        $\Theta$ & parameters of the RFM vector field \\
        $\eta$ & RFM scheduler \\
        $D$ & RFM premetric; mapping $M \times M \mapsto \R^+$ \\
        \bottomrule
    \end{tabular}
\end{table}

%% file: Figures/angles.tex
\tdplotsetmaincoords{70}{290}
\begin{tikzpicture}[every node/.style={font=\footnotesize}]

    \tikzstyle{triangle} = [ultra thin, gray, rounded corners=0.1pt]
    \tikzstyle{geodesic} = [thick, NavyBlue]
    \tikzstyle{point} = [circle, minimum radius=2pt, black]
    \tikzstyle{vector} = [-stealth, thick]
    \tikzstyle{angle} = [draw, very thin, angle radius=4mm]
    
    \def\offx{2.9}
    \def\offy{2.5}
    \def\offytext{4}

    \node (FIG3A) at (0, 0) {
    \begin{tikzpicture}
        \coordinate (P1) at (-1, -1);
        \coordinate (P2) at (-0.6, -1.3);
        \coordinate (P3) at (+0.2, -1.4);
        \coordinate (P4) at (+0.6, -1.1);
        \draw[triangle, fill=black!15] (0, 0) -- (P1) -- (P2) -- cycle;
        \draw[triangle, fill=black!5] (0, 0) -- (P2) -- (P3) -- cycle;
        \draw[triangle, fill=black!10] (0, 0) -- (P3) -- (P4) -- cycle;
        \fill (0, 0) circle (1pt);

    \end{tikzpicture}
    };

    \node (FIG3B) at (\offx,0) {
    \begin{tikzpicture}
        \fill (0, 0) circle (1pt);
        \coordinate (P1) at (-1, 0.8);
        \coordinate (P2) at (0.5, 0.7);
        \coordinate (P3) at (0.8, -0.5);
        \coordinate (P4) at (-0.5, -0.7);
        \draw[triangle, fill=black!25] (0, 0) -- (P1) -- (P2) -- cycle;
        \draw[triangle, fill=black!15] (0, 0) -- (P2) -- (P3) -- cycle;
        \draw[triangle, fill=black!5] (0, 0) -- (P3) -- (P4) -- cycle;
        \draw[triangle, fill=black!15] (0, 0) -- (P1) -- (P4) -- cycle;
        \fill (0, 0) circle (1pt);

    \end{tikzpicture}
    };

    \node (FIG3C) at (2*\offx,0) {
    \begin{tikzpicture}
        \fill (0, 0) circle (1pt);
        \coordinate (P1) at (-1, 0.8);
        \coordinate (P2) at (0.5, -0.1);
        \coordinate (P3) at (0.9, 0.5);
        \coordinate (P4) at (-0.2, -0.7);
        \draw[triangle, fill=black!25] (0, 0) -- (P1) -- (P2) -- cycle;
        \draw[triangle, fill=black!15] (0, 0) -- (P2) -- (P3) -- cycle;
        \draw[triangle, fill=black!5] (0, 0) -- (P3) -- (P4) -- cycle;
        \draw[triangle, fill=black!15] (0, 0) -- (P1) -- (P4) -- cycle;
        \fill (0, 0) circle (1pt);

    \end{tikzpicture}
    };

    \node (FIG2A) at (0, -\offy) {
    \begin{tikzpicture}
        \coordinate (OO) at (0, 0);
        \coordinate (P1) at (-0.9, 0.5);
        \coordinate (P2) at (-1.2, -0.2);
        \coordinate (P3) at (-0.4, -0.9);
        \coordinate (P4) at (+0.9, -0.6);
        \coordinate (P5) at (+0.6, 0.6);
        \draw[triangle, fill=black!15] (OO) -- (P1) -- (P2) -- cycle;
        \draw[triangle, fill=black!5]  (OO) -- (P2) -- (P3) -- cycle;
        \draw[triangle, fill=black!10] (OO) -- (P3) -- (P4) -- cycle;
        \draw[triangle, fill=black!25] (OO) -- (P4) -- (P5) -- cycle;
        \fill (0, 0) circle (1pt);


        \draw pic[angle] {angle = P1--OO--P2};
        \draw pic[angle] {angle = P2--OO--P3};
        \draw pic[angle] {angle = P3--OO--P4};
        \draw pic[angle] {angle = P4--OO--P5};
    \end{tikzpicture}
    };

    \node (FIG2B) at (\offx, -\offy) {
    \begin{tikzpicture}
        \fill (0, 0) circle (1pt);
        \coordinate (P1) at (-1, 0.8);
        \coordinate (P2) at (0.5, 0.7);
        \coordinate (P3) at (0.8, -0.5);
        \coordinate (P4) at (-0.5, -0.7);
        \draw[triangle, fill=black!25] (0, 0) -- (P1) -- (P2) -- cycle;
        \draw[triangle, fill=black!15] (0, 0) -- (P2) -- (P3) -- cycle;
        \draw[triangle, fill=black!5] (0, 0) -- (P3) -- (P4) -- cycle;
        \draw[triangle, fill=black!15] (0, 0) -- (P1) -- (P4) -- cycle;
        \fill (0, 0) circle (1pt);


        \draw pic[angle] {angle = P1--OO--P2};
        \draw pic[angle] {angle = P2--OO--P3};
        \draw pic[angle] {angle = P3--OO--P4};
        \draw pic[angle] {angle = P4--OO--P1};
    \end{tikzpicture}
    };

    \node (FIG2C) at (2*\offx, -\offy) {
    \begin{tikzpicture}
        \coordinate (OO) at (0, 0);
        \coordinate (P1) at (0.7, 1.4);
        \coordinate (P2) at (-1.0, 0.3);
        \coordinate (P3) at (-0.4, -1.0);
        \coordinate (P4) at (+0.9, 0.0);
        \coordinate (P5) at (-0.7, 1.4);
        \draw[triangle, fill=black!25] (OO) -- (P4) -- (P5) -- cycle;
        \draw[triangle, fill=black!15] (OO) -- (P1) -- (P2) -- cycle;
        \draw[triangle, fill=black!5]  (OO) -- (P2) -- (P3) -- cycle;
        \draw[triangle, fill=black!10] (OO) -- (P3) -- (P4) -- cycle;
        \fill (0, 0) circle (1pt);

        \path[name path=line01] (OO) -- (P1);
        \path[name path=line12] (P1) -- (P2);
        \path[name path=line05] (OO) -- (P5);
        \path[name path=line45] (P4) -- (P5);

        \path[name intersections={of=line01 and line45, by=I1}];
        \path[name intersections={of=line12 and line45, by=I2}];
        \draw[dashed, gray, ultra thin] (I1) -- (I2);

        \path[name intersections={of=line12 and line05, by=I3}];
        \draw[dashed, gray, ultra thin] (OO) -- (I3);


        \draw pic[angle] {angle = P1--OO--P2};
        \draw pic[angle] {angle = P2--OO--P3};
        \draw pic[angle] {angle = P3--OO--P4};
        \draw pic[angle, angle radius=5mm] {angle = P4--OO--P1};
        \draw pic[angle, angle radius=5mm, dashed] {angle = P1--OO--P5};

    \end{tikzpicture}
    };

    \draw[-stealth] (FIG3A) -- (FIG2A);
    \draw[-stealth] (FIG3B) -- (FIG2B);
    \draw[-stealth] (FIG3C) -- (FIG2C);

    \node (TEXTA) at (0,       -\offytext) { spherical ($\theta < 2\pi$) };
    \node (TEXTB) at (\offx,   -\offytext) { euclidean ($\theta = 2\pi$) };
    \node (TEXTC) at (2*\offx, -\offytext) { hyperbolic ($\theta > 2\pi$) };
\end{tikzpicture}

%% file: Sections/methodology.tex
\section{Methodology}
\label{sec:method}

\subsection{Discrete differential operators}
In this work, we aim at extending methods defined for Riemannian manifolds to polyhedral meshes.
For this reason, it is necessary to investigate how to properly define the operators described in Sec.~\ref{sec:riemann-manifolds}, namely logarithmic and exponential maps, and parallel transport, in the space of meshes $\mathcal{M}$.

\subsubsection{Tangent cone}
\begin{figure}[bt]
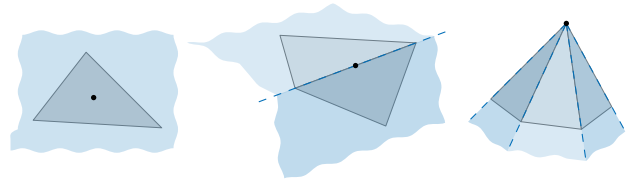

    \centering
    \includestandalone{Figures/tangent-cone} 
    \caption{
    Tangent cone representation (blue-shaded area) for points inside the face (left), on the edge (centre), and on vertices (right).
    }
    \label{fig:tangent-cone}
\end{figure}

On smooth manifolds, it is always possible to define a locally linear approximation of $\M$ at a point $p$, the tangent space $T_p\M$.
Due to the discrete nature of meshes $\mesh$, this concept is ill-defined at edges and vertices.
Still, since meshes are Lipschitz manifolds, we can rely on the natural generalisation of tangent spaces: \textit{tangent cones}.
Following~\cite{Clarke2013}, let $p_i$ be a sequence of points on $\mathcal{M}$ converging to $p$, and $t_i$ a positive sequence decreasing to 0; then the tangent cone $T_p \mathcal{M}$ consists of all vectors $v$ expressible as:
\begin{equation} \label{eq:tangent-cone}
    v = \lim_{i \rightarrow \infty} \frac{p_i - p}{t_i}.
\end{equation}

From this definition, we may identify three cases for tangent cones,
as depicted in Fig.~\ref{fig:tangent-cone}.  If the point $p$ lies in
the interior of a face $\mathcal{T}_i$, then $T_p\mathcal{M}$ is
isomorphic to the 2-dimensional plane containing $\mathcal{T}_i$.  If
$p$ lies on an edge, the tangent cone consists of the union of the two
half-planes defined by the incident faces on the shared edge.
Finally, the tangent cone at a vertex $v_i$ is formed by all
directions spanned by the faces incident to $v_i$, resulting in a
generally non-flat, piecewise-linear cone.

\subsubsection{Logarithmic map}
\label{subsubsec:log_map}
The logarithmic map 
\begin{equation} \label{eq:logmap}
\begin{aligned}
    \text{Log}_p: \, \mathcal{M} & \rightarrow T_p\mathcal{M} \\
    q & \mapsto v
\end{aligned}
\end{equation}
is the operator that maps a point $q$ on the mesh into a vector $v$ in the tangent cone at $p$. The vector $v$ represents the initial direction of the shortest path (geodesic) $\gamma$ from $p$ to $q$, with a magnitude $\|v\|$ equal to the geodesic distance $d_{\mathcal{M}}(p, q)$.

As in the smooth case, this operation requires finding the shortest
path $\gamma$, parameterised by arc-length, joining points $p$ and
$q$.  While this problem can be challenging for arbitrary smooth
manifolds, robust algorithmic solutions for the shortest path problem
exist for polyhedral meshes.  In this respect, the continuous Dijkstra
algorithm~\cite{Mitchell1987} serves as foundation work to compute
discrete geodesics $\gamma$. This method generalises the classical
graph-based Dijkstra algorithm to continuous domains by propagating a
wavefront over the faces of a triangulated surface, maintaining exact
Euclidean distances while respecting surface constraints.  In this
work, we exploit enhanced versions of this
algorithm~\cite{Chen1990,Xin2009} for more efficient
implementations. These methods are based on constructing offline
accelerated query structures that enable the subsequent real--time
generation of shortest paths in fractions of a millisecond.

The shortest path $\gamma$ on $\mesh$ connecting $p$ and $q$ is a polyline which can be easily parameterised by arc-length in the domain $I = [0, \ell]$, with $\ell$ being the path length.
In view of this definition, the vector representation $v$ of the point $q$ in $T_p\mesh$ given by the logarithmic map is defined as
\begin{equation} \label{eq:logmapvalue}
    v = \Log_p(q) \triangleq \ell\,\dot \gamma(0).
\end{equation}
Figure~\ref{fig:logmap} reports a visual representation of this operator.

\begin{figure}
    \centering
    \includestandalone{Figures/logmap}
    \caption{Visual representation of the logarithmic map $v \in T_p\mesh$ of the point $q$; the blue line is the shortest path $\gamma$ on $\mesh$ connecting $p$ and $q$.}
    \label{fig:logmap}
\end{figure}

\subsubsection{Exponential map} \label{sec:expmap}
The exponential map 
\begin{equation} \label{eq:expmap}
\begin{aligned}
    \Exp_p: \, T_p\mesh & \rightarrow \mesh \\
    v & \mapsto q
\end{aligned}
\end{equation}
serves as the inverse operation to the logarithmic map; it maps a vector $v$, defined in the tangent space (or tangent cone) of a point $p$, to a point $q$ on the mesh $\mathcal{M}$.

The exponential map is related to the solution of the initial value problem (IVP) for the geodesic path $\gamma$. Parameterised by arc-length, $\gamma$ starts at $\gamma(0) = p$ with initial velocity $\dot \gamma(0) = v / |v|$ and has an overall length $\ell = |v|$. In smooth manifolds, the solution to this initial value problem is always (locally) unique \cite{doCarmoSurface}, and the geodesic path can be computed via the integration of an Ordinary Differential Equation (ODE) that depends on the intrinsic curvature (and its derivative) of the manifold \cite{Kniely2014}.

In the discrete setting of meshes, the ODE integration approach is
ill-defined, as derivatives cannot be evaluated at edges and vertices,
which constitute the non-differentiable points of the manifold. A
possible strategy to address edge singularities is to compute the
planar unfolding of adjacent faces to propagate the vector
\cite{Mitchell1987}. Despite this improvement, the solution to the
IVP remains ambiguous at vertices, as no unique
shortest path may pass through spherical vertices, and an infinite
number of shortest paths may pass through hyperbolic points. A unified
approach to addressing both vertex and edge singularities was proposed
by Polthier and Schmies \cite{Polthier2006}. In particular, they noted
that smooth geodesics are shortest paths, in the sense that they
minimise length, but are simultaneously straightest paths, in the
sense that they have zero geodesic curvature when measured relative to
the manifold itself. Whilst the former property is insufficient to
uniquely determine the initial value problem at mesh vertices
\cite{Mitchell1987}, the latter consistently yields a unique solution.

\begin{figure}
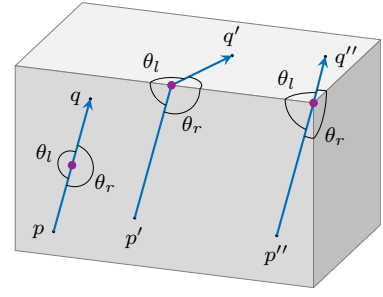

    \centering
    \includestandalone{Figures/straight-geodesic}
    \caption{Straightest geodesics from points $p, p', p''$ to $q, q', q''$. 
    Purple points are points $g$ on which we visualise the left and right curve angles $\theta_l$ and $\theta_r$.}
    \label{fig:straight-geodesic}
\end{figure}

Let $\gamma \subset \mesh$ be a curve on the mesh; given a point $g \in \gamma$, it is possible to compute the left curve angle $\theta_l$ (respectively, right angle $\theta_r$) at the point as the sum of the angles, measured on the mesh, covered clockwise (respectively, counter-clockwise) from the vectors with directions $-\dot \gamma(g)$ and $\dot \gamma(g)$, as shown in Fig.~\ref{fig:straight-geodesic}.
By~\cite[Def. 6]{Polthier2006}, we say that $\gamma$ is a straightest geodesic if, for any point $g$ in the curve, the left and right curve angles $\theta_l$ and $\theta_r$ are equal, and evaluate to half the total angle~\eqref{eq:totalangle}.

\input{algorithms/exp-map.tex}

\begin{figure}
    \centering
    \includestandalone{Figures/exp-map-example} \vspace{-7mm}
    \caption{Visual representation of the logarithmic map $v \in T_p\mesh$ of the point $q$; the blue line is the shortest path $\gamma$ on $\mesh$ connecting $p$ and $q$.}
    \label{fig:expmap-example}
\end{figure}

Exploiting this definition, we define a numerical procedure---outlined in Algorithm~\ref{alg:exp-map}---to compute the exponential map~\eqref{eq:expmap} that is always well-defined, even when $v$ passes through vertices of $\mesh$.  
In a iterative fashion, the algorithm displaces $p$ along the direction $v$ and uses the straightest geodesic definition to propagate motion over non-differentiable points for the manifold.
Upon initialisation (line 1), we compute the tentative displaced point $\hat p_{k+1}$ (line 3).
If such a point lies in the interior of the face $\face_k$ containing $p_k$ and $v_k$ (line 4--5), then we completed the straightest geodesic propagation and $\hat p_{k+1}$ represents the final exponential mapping $q$ (line 6--7); this case is visually depicted in Fig.~\ref{fig:expmap-example}(a).
Otherwise, Fig.~\ref{fig:expmap-example}(b), when $\hat p_{k+1}$ falls outside $\face_k$, we apply straightest geodesic propagation at the singular point; 
let $p_{k+1}$ be the point in which $v_k$ intersects its containing face (line 9), we compute its total angle $\theta$ (line 10) and the remainder vector $w$ (line 11) which accounts for the motion between $p_k$ and $p_{k+1}$; by rotating $-w$ intrinsically on the mesh by $\theta / 2$ (line 12), we construct the vector $v_{k+1}$ generating the straightest path at $p_{k+1}$.
By iterating this procedure, we ensure that $v_k$ constantly decreases in length, finally reaching the terminal projection point $q$.

\subsubsection{Parallel transport} 
The parallel transport
\begin{equation} \label{eq:partransp}
\begin{aligned}
    P_{p\rightarrow q}: \, T_p\mesh & \rightarrow T_q\mesh \\
    v & \mapsto w
\end{aligned}
\end{equation}
is the operator that enables the expression of a vector $v$, defined in the tangent cone of a point $p$, into the different tangent cone placed at $q$.
More in detail, $P_{p\rightarrow q}$ defines an isometric transformation that moves vector $v$ into $w$ with zero covariant derivative along the path connecting $p$ and $q$.

\begin{figure}
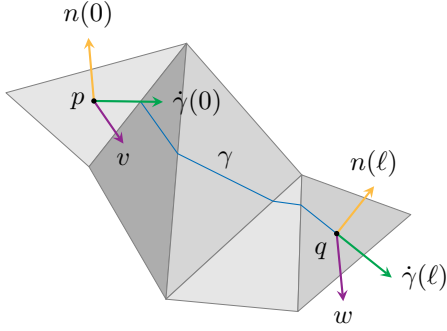

    \centering
    \includestandalone{Figures/partransp}
    \caption{Visual representation of the parallel transport of the vector $v \in T_p\mesh$ into the the tangent cone of $q$, yielding $w \in T_q\mesh$.
    We observe that the angle between and $\dot \gamma(0)$ and $v$ is equal to the angle between $\dot \gamma(\ell)$ and $w$.}
    \label{fig:partransp}
\end{figure}

In our case, considering that the mesh $\mesh$ is embedded in the Euclidean space $\R^3$, it follows that the isometric transformation is a rotation, that we can encode as a matrix $R \in \R^{3\times 3}$.
To construct such a rotation, we first need to solve the boundary value problem of computing the geodesic path $\gamma$ connecting $p$ and $q$, as discussed in Sec.~\ref{subsubsec:log_map}.
Since the velocity vector field is parallel to itself when moving along a geodesic, the parallel transport transformation~\eqref{eq:partransp} can be evaluated as
\begin{align}
    R(\ell) &  \triangleq \begin{bmatrix} 
        \dot{\gamma}(\ell) & n(\ell) \times \dot \gamma(\ell) & n(\ell)
    \end{bmatrix},  \\
    w & \triangleq R(\ell) R^\top(0) \, v \in T_q\mesh,
\end{align}
where $n(\cdot) : I \rightarrow \R^3$ is function that assigns each point on the geodesic $\gamma$ its surface normal vector; this operator is visually represented in Fig.~\ref{fig:partransp}.

\subsubsection{Ambient space and projections}
\label{sec:projs}

Most learning methods are based on learning some vector fields $\psi: \mesh \rightarrow T_{(\cdot)}\mesh$ defined in the manifold itself.
The main problem of this formulation is that it hardly integrates with existing, state of the art implementations which are defined in Euclidean spaces.

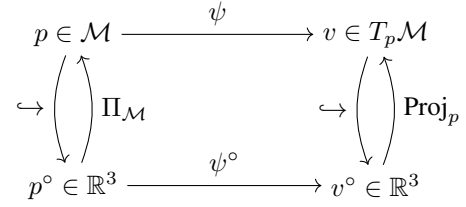
\begin{figure}
    \centering
    \begin{tikzpicture}
        \node (ym) at (0, 0)  {$p \in \mesh$};
        \node (ya) at (0, -2) {$p^\circ \in \R^3$};
        \node (vm) at (4, 0)  {$v \in T_p\mesh$};
        \node (va) at (4, -2) {$v^\circ \in \R^3$};
    
        \draw [->] (ym) to[bend right=20] node[midway, left]{$\hookrightarrow$} (ya);
        \draw [->] (ya) to[bend right=20] node[midway, right]{$\Pi_\mesh$} (ym);
        \draw [->] (vm) to[bend right=20] node[midway, left]{$\hookrightarrow$} (va);
        \draw [->] (va) to[bend right=20] node[midway, right]{$\textrm{Proj}_p$} (vm);
        \draw [->] (ym) -- (vm) node[midway, above]{$\psi$};
        \draw [->] (ya) -- (va) node[midway, above]{$\psi^\circ$};
    \end{tikzpicture}
    \caption{Relationship between elements defined in the mesh domain and in the ambient space.}
    \label{fig:conversions}
\end{figure}

For this reason, being able to work in the ambient space $\R^3$ and constructing an "augmented" vector field $\psi^\circ: \R^3\rightarrow \R^3$ approximating $\psi$ on $\mesh$ is often a simpler approach.
Based on this idea, it is necessary to define operations to project points and vectors back and forth between the mesh and its embedding space, as depicted in Fig.~\ref{fig:conversions}.

While transforming a point $p$ from the mesh to $p^\circ$ on Euclidean
space is trivial, the inverse operation is not.  The simplest approach
is, in this case, to use the closest point function $\Pi_\mesh$ to
project point back to the discrete manifold:
\begin{equation} \label{eq:point-projection}
    \Pi_\mesh(p^\circ) \triangleq \arg \min_{p \in \mesh} \| p - p^\circ \|.
\end{equation}
In practice, this projection can be efficiently computed using suitable data structure---such as Axis-Aligned Bounding-Boxes~\cite{Ericson2004}---that accelerate closest-point queries.

In the same way, for vectors in the tangent cone $T_p\mesh$, which are
trivially embedded in $\R^3$, we define $\textrm{Proj}_p$ as the
projection function that maps vector in $\R^3$ back to the tangent
cone $T_p\mesh$ at the point $p$:
\begin{equation} \label{eq:vector-projection}
    \textrm{Proj}_p(v^\circ) \triangleq \arg \min_{v \in T_p\mesh} \| v - v^\circ \|.
\end{equation}
If $p$ lies in the interior of a face, then $T_p\mesh$ becomes a plane
and $\textrm{Proj}_p(v^\circ)$ simplifies to the normal projection to
the plane itself.  In all other cases, we rely on the numerical
procedure described in Algorithm~\ref{alg:projection} to perform the
projection: first, we project the vector $v$ normally onto all faces
(lines 2--8) and onto directed edges (lines 9--14) incident to the
point $p$, and remove all projections outside the tangent cone
$T_p\mesh$; we then select (line 15) as projection the candidate which
maximises the dot-product w.r.t. the initial vector $v$.

\input{algorithms/tv-proj.tex}

\subsection{G-DMPs on Meshes}

From the above formulation of the operators on the mesh manifold, it
is straightforward to generalise the geometry-aware DMP formulation
in~\eqref{eq:gdmp} to the space of discrete meshes.  The core elements
that enables the learning of the actual demonstration in DMP is the
forcing term $f$, a function that shall map the phase variable in $\R$
into the tangent cone $T_pM$ of the current position state $p$.  A
standard way of learning such forcing term is by considering the
tangent bundle as an inclusion of the ambient space of the manifold
itself; in this way, the DMP effectively learns a forcing term living
in a bigger space dimension, e.g., $\R^3$ in the case of surfaces, and
then either a) they project $f$ in $T_pM$ before integrating the
systems dynamic, or b) they integrate the unconstrained dynamical
system and perform projection of the position state $y$ once it starts
deviating too much from the manifold itself.

We believe that such approach yields poor generalisation capability:
considering a policy learned in the trivial surface given by the $xy$
plane, then any attempt in integrating the same forcing term on the
$xz$ plane would not produce meaningful results. To overcome this
issue, other works~\cite{Xue2023, DalleVedove25} defined the forcing
term $f$ to be locally parameterised by the current state $p$; while
this solution overcomes the problem of having transferable motion
policies, the locality of the parameterisation may fail in capturing
global behaviour of the motion policy itself.

In this work, we propose to learn the forcing term $f$ in the tangent cone of a fixed point, and then use parallel transport when executing the motion to retrieve a value properly defined in $T_p\mesh$; for this scope, we select as fixed point the goal configuration $g$.
Given a demonstration $\mathcal D = \left\{p_k, \dot p_k, \ddot p_k\right\}_{k=1}^D$, with $p_k \in \mesh$ and $\dot p_k, \ddot p_k \in T_p \mesh$, firstly we compute the local forcing term $f_{p,\cdot}$ by inverting~\eqref{eq:gdmp}, and then use parallel transport to compute the forcing term $f_{g,\cdot}$ at the motions goal:
\begin{align}
    f_{p,k} & = \tau^2 \ddot p_k - \alpha \big( \beta \Log_{p_k}(g) - \tau \dot p_k\big), \\
    f_{g,k} & = P_{p_k \rightarrow g}\left(f_{p,k}\right).
\end{align}
In this way, the function $f_{g,k}$ lies in $T_g\mesh$ for all $D$ samples; upon the definition of an isomorphism from the bi-dimensional subspace in $\R^3$ given by $T_g\mesh$ to $\R^2$, the forcing term can be encoded in a minimal representation as the linear combination of $N_b$ non-linear basis functions.

To generalise the motion policy to other final points or different surfaces, it is sufficient to select a new $g' \in \mesh'$ in a potentially different mesh $\mesh'$, and choose a suitable parameterisation to reconstruct a forcing term $f \in T_{g'}\mesh'$ from the encoded 2D representation.
While integrating, $f(x)$ is first evaluated at $T_{g'}\mesh'$ and then parallel transported to the current position $p$ tangent cone $T_p\mesh'$, effectively yielding the following manifold-ODE:
\begin{equation} \label{eq:newmeshdmp}
    \begin{aligned}
        \tau \nabla_z z & = \alpha ( \beta \Log_p(g) - z) + P_{g \rightarrow p}\big(f(x)\big),  \\
        \tau \dot p & = z, \\
        \tau \dot x & = -\zeta x.
    \end{aligned}
\end{equation}

While~\eqref{eq:newmeshdmp} can be used to encode and generalise discrete motions, we can use the rhythmic formulation of DMPs to enable learning and execution of periodic motion:
\begin{equation} \label{eq:rhythdmp}
    \begin{aligned}
        \nabla_z z & = \Omega \Big(\alpha ( \beta \Log_p(g) - z) + P_{g \rightarrow p}\big(f(\phi)\big) \Big),  \\
        \dot p     & = \Omega z, \\
        \dot \phi  & = 2\pi \Omega,
    \end{aligned}
\end{equation}
where in this case $g$ becomes the centre of the motion and $\Omega$ is a time scaling factor associated to the frequency of execution of the motion policy.

\subsection{Geodesic-based Gaussian Processes}

In this work, we are interested in evaluating GPs~\eqref{eq:gp} for which the input space $\mathcal X$ is the mesh $\mesh$.
When working on manifolds, the main challenge of using GPs is the definition of a valid kernel function $k$ that must be positive definite for any choice of points in the manifold itself.
In Euclidean domain, the squared exponential kernel
\begin{equation} \label{eq:euclidean-sek}
    k(x, x') = \sigma^2 \exp\left(- \frac{\|x-x'\|^2}{2\kappa^2} \right)
\end{equation}
is one of the most popular choices to induce infinitely mean-square differentiable GPs, with the variance $\sigma^2$ and length scale $\kappa$ as hyperparameters.

In this work, we propose to use a kernel in which  the norm $\|\cdot\|$ in the Euclidean space is simply replaced with the geodesic distance $d_g$ on the manifold itself:
\begin{equation} \label{eq:mesh-kernel}
    k_d(x, x') = \sigma^2 \exp\left(- \frac{d_g^2(x, x')}{2\kappa^2} \right).
\end{equation}
While~\cite[Theorem. 2]{Feragen2015} proves that such a kernel may yield negative-definite covariance matrices for some $\kappa > \kappa_{th}$, we believe that in most practical application the threshold $\kappa_{th}$ is big enough to guarantee a correct GPR.

\input{algorithms/gp-threshold.tex}

For this reason, we designed Algorithm~\ref{alg:kernel-bisection} to
estimate the value of the maximum admissible length scale value given
the set of training data for the GP.  We first compute the $m\times m$
distance matrix $D$ between the a set of input points (lines 1--2);
then, we follow a bisection based approach to find the critical
length scale $\kappa_{th}$: let $\kappa_{min}$ (respectively,
$\kappa_{max}$) be a length scale for which the kernel is positive
(respectively, negative) definite, we first initialise
$\kappa_{min} = \kappa_0$ and $\kappa_{max} = 2\kappa_0$ and iterate
(lines 3--9) until we find the pair of length scales satisfying our
assumptions; we proceed (lines 10--17) using a standard bisection
algorithm when we iteratively evaluate and update the search range
based on the positive definiteness of the kernel at the length scale
$\kappa_{mid}$ in the middle of $\kappa_{min}$ and $\kappa_{max}$,
until numerical convergence.  Algorithm~\ref{alg:kernel-bisection} can
be used to compute the critical length scale $\kappa_{th}$ when the set
$P$ of test point is known; if such set is not available upfront, it
is also possible to run a Montecarlo identification of $\kappa_{th}$
by randomly sampling points on $\mesh$, computing the critical
length scale with the algorithm, and choosing the minimum $\kappa_{th}$
among all tests.

\subsection{Riemannian Flow Matching on Meshes}

To successfully implement RFM on meshes, it is necessary to choose a valid premetric $D$ for the space of discrete surfaces. 
For general geometries, original work~\cite{Chen2023} proposes the use of spectral-based distances---which can be computed by the Laplace-Beltrami operator---and use a ODE solver on that metric to compute the reference vector field of the training objective~\eqref{eq:rcfm}.

In this work, we propose to directly use the shortest path on $\mesh$ to measure distance between samples, effectively choosing the geodesic distance $d_g$ as valid premetric. 
Let $p_0 \backsim \pi_\textrm{ref}$ and $p_1 \backsim \pi_\textrm{data}$ two samples on the mesh drawn respectively from the reference and data distributions, we select as corresponding flow $\psi_t$ the shortest path $\gamma: [0, 1] \rightarrow \mesh$ where $\gamma(0) = p_0$ and $\gamma(1) = p_1$; 
once $\gamma$ is constructed, sampling points $p_t$ along the flow for times $t \in \mathcal U(0, 1)$, as well as the corresponding tangential velocity vector $u_t \propto \dot \gamma(t)$, becomes trivial.
Algorithm~\ref{alg:rfm} shows how to train RFM on meshes, using the linear scheduler $\eta(t) = 1 - t$ as in~\cite{Chen2023}.

\input{algorithms/rfm.tex}

To avoid restrictions on the topology of the mesh and learn a parameterisation-free vector field, the NN describing $v_t$ directly works in the ambient space $\R^3$ of the mesh. 
Point from the reference distribution $\pi_\textrm{ref}$ are uniformly sampled by first selecting a mesh face with probability proportional to its area, and then drawing a point uniformly within the interior of that face.

To drawn a sample distributed as $\pi_\textrm{data}$ given the trained vector field $v_t$, we first extract a sample $p_0$ from the reference distribution and perform a mesh-based integration technique over the field by consistently projecting the NN outcome on the mesh tangent cone using~\eqref{eq:vector-projection}, followed by the geometrically-consistent exponential map operation~\eqref{eq:expmap}, as shown in Algorithm~\ref{alg:rfm-sampling}.

\input{algorithms/rfm-sampling.tex}

%% file: Figures/tangent-cone.tex
\begin{tikzpicture}[every node/.style={font=\footnotesize}]

    \tikzstyle{triangle} = [ultra thin, gray, rounded corners=0.1pt]
    \tikzstyle{geodesic} = [thick, NavyBlue]
    \tikzstyle{point} = [circle, minimum radius=2pt, black]
    \tikzstyle{vector} = [-stealth, thick]
    \tikzstyle{angle} = [draw, very thin, angle radius=4mm]
    \tikzstyle{boarder} = [decoration={snake, amplitude=0.3mm, segment length=5mm}]
    \tikzstyle{tangentcone} = [fill=NavyBlue, opacity=0.2]
    \tikzstyle{tangentedge} = [dashed, NavyBlue]

    \def\offx{3.0}

    \node (FIG3A) at (0, 0) {
    \begin{tikzpicture}
        \coordinate (P1) at (-1, -0.4);
        \coordinate (P2) at (0.7, -0.5);
        \coordinate (P3) at (-0.3, 0.5);
        \draw[triangle, fill=black!15] (P1) -- (P2) -- (P3) -- cycle;

        \def\dx{1.0}
        \def\dy{0.8}
        \coordinate (CC) at (-0.3, -0.0);
        \coordinate (E1) at ($(CC) + (-\dx, -\dy)$);
        \coordinate (E2) at ($(CC) + (\dx, -\dy)$);
        \coordinate (E3) at ($(CC) + (\dx, \dy)$);
        \coordinate (E4) at ($(CC) + (-\dx, \dy)$);

        \fill[boarder][tangentcone] decorate { (E1) -- (E2) -- (E3) -- (E4) -- cycle };

        \fill (-0.2, -0.1) circle (1pt);
    \end{tikzpicture}
    };

    \node (FIG3B) at (\offx,0) {
    \begin{tikzpicture}

        \coordinate (D1) at (0.8, 0.3);

        \coordinate (OO) at (0, 0);
        \coordinate (P1) at ($(D1)$);
        \coordinate (P2) at ($-1.0*(D1)$);
        \coordinate (P3) at (-1.0, 0.4);
        \coordinate (P4) at (0.4, -0.8);
        \draw[triangle, fill=black!7] (P1) -- (P2) -- (P3) -- cycle;
        \draw[triangle, fill=black!15] (P1) -- (P2) -- (P4) -- cycle;

        \draw[tangentedge] ($-1.6*(D1)$) -- ($1.6*(D1)$);

        \coordinate (E1) at ($1.3*(P1)$);
        \coordinate (E2) at ($1.3*(P2)$);
        \coordinate (T1) at ($(E1) + 1.11*(-1.0, 0.3)$);
        \coordinate (T2) at ($(E2) + 1.11*(-1.0, 0.5)$);
        \coordinate (Y1) at ($(E1) + 1.11*(0.1, -1.0)$);
        \coordinate (Y2) at ($(E2) + 1.11*(0.1, -1.0)$);
        \fill[boarder][tangentcone, opacity=0.15] (E2) -- (E1) decorate { (E1) -- (T1) -- (T2) -- (E2) };
        \fill[boarder][tangentcone, opacity=0.25] (E2) -- (E1) decorate { (E1) -- (Y1) -- (Y2) -- (E2) };

        \fill (0, 0) circle (1pt);
    \end{tikzpicture}
    };

    \node (FIG3C) at (2*\offx,0) {
    \begin{tikzpicture}
        \coordinate (OO) at (0, 0);
        \coordinate (P1) at (-1, -1);
        \coordinate (P2) at (-0.6, -1.3);
        \coordinate (P3) at (+0.2, -1.4);
        \coordinate (P4) at (+0.6, -1.1);
        \draw[triangle, fill=black!15] (0, 0) -- (P1) -- (P2) -- cycle;
        \draw[triangle, fill=black!5] (0, 0) -- (P2) -- (P3) -- cycle;
        \draw[triangle, fill=black!10] (0, 0) -- (P3) -- (P4) -- cycle;

        \coordinate (E1) at ($1.3*(P1)$);
        \coordinate (E2) at ($1.3*(P2)$);
        \coordinate (E3) at ($1.3*(P3)$);
        \coordinate (E4) at ($1.3*(P4)$);

        \fill[boarder][tangentcone, opacity=0.25] (E2) -- (OO) -- (E1) decorate { (E1) -- (E2) };
        \fill[boarder][tangentcone, opacity=0.15] (E3) -- (OO) -- (E2) decorate { (E2) -- (E3) };
        \fill[boarder][tangentcone, opacity=0.25] (E4) -- (OO) -- (E3) decorate { (E3) -- (E4) };
        \draw[tangentedge] (OO) -- (E1);
        \draw[tangentedge] (OO) -- (E2);
        \draw[tangentedge] (OO) -- (E3);
        \draw[tangentedge] (OO) -- (E4);

        \fill (0, 0) circle (1pt);
    \end{tikzpicture}
    };

\end{tikzpicture}

%% file: Figures/logmap.tex
\begin{tikzpicture}

    \tikzstyle{triangle} = [ultra thin, gray, rounded corners=0.1pt]
    \tikzstyle{geodesic} = [NavyBlue]
    \tikzstyle{point} = [circle, minimum radius=2pt, black]
    \tikzstyle{vector} = [-stealth, thick]

    \coordinate (V0) at (-2.396906868625721, 1.0481795585910636);
    \coordinate (V1) at (-1.2975402707849137, 0.06910588054655342);
    \coordinate (V2) at (-0.006646638672219152, 1.7013665275540721);
    \coordinate (V3) at (-0.2799599922189223, -1.6825249405893197);
    \coordinate (V4) at (1.4648534216782325, -1.8454682951300705);
    \coordinate (V5) at (1.5170648673289517, -0.03875338944440654);
    \coordinate (V6) at (2.9592631095041253, -0.5635338123497262);
    \draw[triangle, fill=black!10] (V0) -- (V1) -- (V2) -- cycle;
    \draw[triangle, fill=black!32] (V1) -- (V3) -- (V2) -- cycle;
    \draw[triangle, fill=black!16] (V3) -- (V5) -- (V2) -- cycle;
    \draw[triangle, fill=black!10] (V4) -- (V5) -- (V3) -- cycle;
    \draw[triangle, fill=black!18] (V6) -- (V5) -- (V4) -- cycle;

    \fill[opacity=0.13, Dandelion] (3.9034533924541845, 1.5114943429650176) -- (2.3543810339189513, -0.44721843344400525) -- (-3.4035032319416594, 0.09049463654047218) -- (-1.8544308734064263, 2.049207412949495) -- cycle; 
    \node[right, black, opacity=100] at  (3.9034533924541845, 1.5114943429650176) {$T_p\mathcal M$};

    \draw[vector, Plum] (-1.233116321556319, 0.9394963411123829) -- (2.650466055226451, 0.8872375141985944) node[right, black] {$v$};

    \coordinate (P0) at (-1.233116321556319, 0.9394963411123829);
    \coordinate (P1) at (-0.6157512229614374, 0.9311888629388901);
    \coordinate (P2) at (-0.12447174792832878, 0.24257435084221546);
    \coordinate (P3) at (1.1368229471533944, -0.38656763787642445);
    \coordinate (P4) at (1.505564637159299, -0.43670516050480734);
    \coordinate (P5) at (1.9787739704239937, -0.8147908959521499);
    \draw[geodesic] (P0) -- (P1) -- (P2) -- (P3) -- (P4) -- (P5);
    \node[above] at (0.5061755996125328, -0.07199664351710455) {$\gamma$};

    \fill[black] (-1.233116321556319, 0.9394963411123829) circle (1pt) node[above, black] {$p$};

    \fill[black] (1.9787739704239937, -0.8147908959521499) circle (1pt) node[above right, black] {$q$};

\end{tikzpicture}

%% file: Figures/straight-geodesic.tex
\tdplotsetmaincoords{70}{290}
\begin{tikzpicture}[tdplot_main_coords, scale=2, every node/.style={font=\footnotesize}]

    \tikzstyle{triangle} = [ultra thin, gray, rounded corners=0.1pt]
    \tikzstyle{geodesic} = [thick, NavyBlue]
    \tikzstyle{point} = [circle, minimum radius=2pt, black]
    \tikzstyle{vector} = [-stealth, thick]
    \tikzstyle{angle} = [very thin]
    \def\ptrad{0.8}  
    \def\ptpq{0.3}  

    \def\dx{1.3} 
    \def\dy{2.1} 
    \def\dz{1.3} 

    \def\vang{15} 
    \def\phi{15}  
    \def\lf{1.0}

    \def\offy{1.0}

    \def\cy{1.7};
    \def\cz{-{0.5*\dz}};

    \coordinate (DPO1) at ($(0, {sin(\vang)}, {-cos(\vang)})$);
    \coordinate (DOQ1) at ($({cos(\phi)}, {sin(\phi)}, 0)$);
    \coordinate (DPO2) at ($(0, {sin(\vang)}, {-cos(\vang)})$);
    \coordinate (DOQ2) at ($({cos(\vang)}, -{sin(\vang)}, 0)$);
    \coordinate (O1) at (0, 0, 0);
    \coordinate (O2) at (0, \offy, 0);
    \coordinate (O3) at ($(0, \cy, \cz)$);

    \coordinate (P1) at ($(O1) + \lf*(DPO1)$);
    \coordinate (P2) at ($(O2) + \lf*(DPO2)$);
    \coordinate (Q1) at ($(O1) + 0.9*(DOQ1)$);
    \coordinate (Q2) at ($(O2) + 0.7*(DOQ2)$);

    \coordinate (P3) at ($(O3) + 0.5*(DPO1)$);
    \coordinate (Q3) at ($(O3) - 0.5*(DPO1)$);

    \draw[triangle, fill=black!5] (0, 0, 0) -- (\dx, 0, 0) -- (\dx, \dy, 0) -- (0, \dy, 0) -- cycle;
    \draw[triangle, fill=black!15] (0, 0, 0) -- (0, \dy, 0) -- (0, \dy, -\dz) -- (0, 0, -\dz) -- cycle;
    \draw[triangle, fill=black!25] (0, 0, 0) -- (\dx, 0, 0) -- (\dx, 0, -\dz) -- (0, 0, -\dz) -- cycle;

    \fill[black] (P1) circle (\ptpq pt) node[below]{$p''$};
    \fill[black] (P2) circle (\ptpq pt) node[below]{$p'$};
    \fill[black] (P3) circle (\ptpq pt) node[left]{$p$};
    \fill[black] (Q1) circle (\ptpq pt) node[right]{$q''$};
    \fill[black] (Q2) circle (\ptpq pt) node[above]{$q'$};
    \fill[black] (Q3) circle (\ptpq pt) node[left]{$q$};

    \draw[geodesic, -stealth] (P1) -- (O1) -- (Q1);
    \draw[geodesic, -stealth] (P2) -- (O2) -- (Q2);
    \draw[geodesic, -stealth] (P3) -- (O3) -- (Q3);

    \def\rr{0.25}  
    \def\rl{0.2}  
    \draw[canvas is yz plane at x=0, angle]
        (0,0) ++({-90+\vang}:\rr) arc[start angle={-90+\vang}, end angle=-90, radius=\rr];
    \draw[canvas is xz plane at y=0, angle]
        (0,-\rr) arc[start angle=-90, end angle=0, radius=\rr] node[pos=0.1, right]{$\theta_r$};
    \draw[canvas is xy plane at z=0, angle]
        (\rr, 0) arc[start angle=0, end angle=\phi, radius=\rr];

    \draw[canvas is yz plane at x=0, angle]
        (0,0) ++({-90+\vang}:\rl) arc[start angle={-90+\vang}, end angle=0, radius=\rl];
    \draw[canvas is xy plane at z=0, angle]
        (0, \rl) arc[start angle=90, end angle=\phi, radius=\rl] node[pos=0.1, above]{$\theta_l$};

    \def\rr{0.2}  
    \def\rl{0.15}  
    \draw[canvas is yz plane at x=0, angle]
        (\offy, 0) ++({-90+\vang}:\rr) arc[start angle={-90+\vang}, end angle=-180, radius=\rr] node[pos=0.6, below]{$\theta_r$};
    \draw[canvas is xy plane at z=0, angle]
        (0, \offy) ++ (0, -\rr) arc[start angle=-90, end angle=-\vang, radius=\rr];

    \draw[canvas is yz plane at x=0, angle]
        (\offy, 0) ++({-90+\vang}:\rl) arc[start angle={-90+\vang}, end angle=0, radius=\rl];
    \draw[canvas is xy plane at z=0, angle]
        (0, \offy) ++ (0, \rl) arc[start angle=90, end angle=-\vang, radius=\rl] node[pos=0.1, above]{$\theta_l$};

    \def\rr{0.15}  
    \def\rl{0.1}  

    \draw[canvas is yz plane at x=0, angle]
        (\cy, \cz) ++({-90+\vang}:\rr) arc[start angle={-90+\vang}, end angle={-270+\vang}, radius=\rr]
        node [pos=0.3, right]{$\theta_r$};

    \draw[canvas is yz plane at x=0, angle]
        (\cy, \cz) ++({-90+\vang}:\rl) arc[start angle={-90+\vang}, end angle={90+\vang}, radius=\rl]
        node [pos=0.7, left]{$\theta_l$};

    \fill[Plum] (O1) circle (\ptrad pt);
    \fill[Plum] (O2) circle (\ptrad pt);
    \fill[Plum] (O3) circle (\ptrad pt);


\end{tikzpicture}

%% file: algorithms/exp-map.tex
 \begin{algorithm}[bt]
     \caption{Iterative procedure to compute the exponential map~\eqref{eq:expmap}. Figure~\ref{fig:expmap-example} visually shows the quantities of interested when $\hat p_{k+1}$ lies either on the inside or outside of the face $\face_k$.}
    \label{alg:exp-map}
 \begin{algorithmic}[1]
    \renewcommand{\algorithmicrequire}{\textbf{Input:}}
    \Require $p \in \mesh$ application point, $v \in T_p\mesh$ vector

    \renewcommand{\algorithmicensure}{\textbf{Output:}}
    \Ensure $q \in \mesh$

    \State $p_0 \gets p,\ v_0 \gets v$

    \While{$|v_k| > 0$}
        \State $\hat p_{k+1} \gets p_k + v_k$
        \State $\face_k \gets$ face containing $p_k, v_k$
        \If{$\hat p_{k+1} \in \face_k$}
            \State $q \gets \hat p_{k+1}$
            \State Return
        \EndIf
        \State $p_{k+1} \gets \text{intersection}\left( \overline{p_k \hat p_{k+1}}, \text{boundary of } \face_k \right) $
        \State $\theta \gets $ total angle~\eqref{eq:totalangle} at $p_{k+1}$
        \State $w \gets v_k - (p_{k+1} - p_k)$
        \State $v_{k+1} \gets $ rotate $-w$ on $\mesh$ w.r.t. $p_{k+1}$ by $\theta / 2$
        \State $k \gets k + 1$
    \EndWhile
\end{algorithmic}
\end{algorithm}

%% file: Figures/exp-map-example.tex
\tdplotsetmaincoords{70}{290}
\begin{tikzpicture}[every node/.style={font=\footnotesize}]
    \tikzstyle{triangle} = [ultra thin, gray, rounded corners=0.1pt]
    \tikzstyle{geodesic} = [thick, NavyBlue]
    \tikzstyle{point} = [circle, minimum radius=2pt, black]
    \tikzstyle{vector} = [-stealth, thick]
    \tikzstyle{angle} = [draw, very thin, angle radius=4mm]
    \tikzstyle{boarder} = [decoration={snake, amplitude=0.3mm, segment length=5mm}]
    \tikzstyle{tangentcone} = [fill=NavyBlue, opacity=0.2]
    \tikzstyle{tangentedge} = [dashed, NavyBlue]

    \def\offx{4.5}
    \def\offy{1.7}

    \node (FIG3A) at (0, 0) {
    \begin{tikzpicture}
        \coordinate (P1) at (-1, -0.6);
        \coordinate (P2) at (0.9, -0.8);
        \coordinate (P3) at (-0.2, 0.8);
        \coordinate (Pk) at ($(P1)!0.3!(P2)$);
        \coordinate (Vk) at ($1.6*(0.2, 0.6)$);
        \coordinate (Q) at ($(Pk) + (Vk)$);
        \draw[triangle, fill=black!15] (P1) -- (P2) -- (P3) -- cycle;

        \def\dx{1.0}
        \def\dy{0.8}

        \fill (Pk) circle (1pt) node[below]{$p_k$};
        \fill (Q) circle (1pt) node[above right]{$\hat p_{k+1} = q$};
        \draw[vector, NavyBlue] (Pk) -- (Q) node[midway, left, black]{$v_k$};
        \node[below]  at ($(P1)!0.8!(P2)$) {$\mathcal T_k$};
    \end{tikzpicture}
    };

    \node (FIG3B) at (\offx,0) {
    \begin{tikzpicture}[tdplot_main_coords, scale=2, every node/.style={font=\footnotesize}]

        \coordinate (A) at (-1, 0, 0);
        \coordinate (B) at (1.3, 0, 0);
        \coordinate (C) at (0.6, -0.8, -0.8);
        \coordinate (D) at (0.2, 1, 0);

        \coordinate (P) at (-0.04999999999999999, 0.6, 0);
        \coordinate (Vk) at (0.72, -1.56, 0);
        \coordinate (Phat) at ($(P) + (Vk)$);
        \coordinate (Pnext) at  (0.2269230769230769, 0, 0);
        \coordinate (Q) at  (0.6700000000000002, -0.6788225099390857, -0.6788225099390857);

        \coordinate (N) at (0, -0.7071067811865476, 0.7071067811865476);
        \coordinate (Wneg) at (-0.2161538461538463, 0.9600000000000001, 0);

        \draw[triangle, fill=black!10] (A) -- (B) -- (D) -- cycle;
        \draw[triangle, fill=black!20] (A) -- (B) -- (C) -- cycle;

                \draw[dashed] (-0.216, 0.960, 0.000) -- (-0.253, 0.942, 0.000) -- (-0.290, 0.922, 0.000) -- (-0.326, 0.901, 0.000) -- (-0.361, 0.879, 0.000) -- (-0.395, 0.855, 0.000) -- (-0.428, 0.830, 0.000) -- (-0.460, 0.804, 0.000) -- (-0.491, 0.776, 0.000) -- (-0.521, 0.747, 0.000) -- (-0.550, 0.717, 0.000) -- (-0.577, 0.686, 0.000) -- (-0.604, 0.654, 0.000) -- (-0.629, 0.621, 0.000) -- (-0.652, 0.587, 0.000) -- (-0.675, 0.552, 0.000) -- (-0.696, 0.516, 0.000) -- (-0.715, 0.480, 0.000) -- (-0.733, 0.442, 0.000) -- (-0.750, 0.404, 0.000) -- (-0.765, 0.366, 0.000) -- (-0.779, 0.327, 0.000) -- (-0.791, 0.287, 0.000) -- (-0.801, 0.247, 0.000) -- (-0.810, 0.206, 0.000) -- (-0.817, 0.165, 0.000) -- (-0.823, 0.124, 0.000) -- (-0.827, 0.083, 0.000) -- (-0.830, 0.041, 0.000) -- (-0.830, 0.000, 0.000) -- (-0.830, 0.000, 0.000) -- (-0.828, -0.052, -0.052) -- (-0.820, -0.103, -0.103) -- (-0.808, -0.154, -0.154) -- (-0.790, -0.204, -0.204) -- (-0.768, -0.253, -0.253) -- (-0.741, -0.301, -0.301) -- (-0.709, -0.348, -0.348) -- (-0.673, -0.392, -0.392) -- (-0.633, -0.435, -0.435) -- (-0.588, -0.476, -0.476) -- (-0.540, -0.515, -0.515) -- (-0.487, -0.551, -0.551) -- (-0.432, -0.585, -0.585) -- (-0.373, -0.615, -0.615) -- (-0.312, -0.643, -0.643) -- (-0.248, -0.668, -0.668) -- (-0.181, -0.690, -0.690) -- (-0.113, -0.708, -0.708) -- (-0.043, -0.723, -0.723) -- (0.028, -0.734, -0.734) -- (0.100, -0.742, -0.742) -- (0.173, -0.747, -0.747) -- (0.246, -0.748, -0.748) -- (0.319, -0.745, -0.745) -- (0.391, -0.739, -0.739) -- (0.463, -0.729, -0.729) -- (0.534, -0.716, -0.716) -- (0.603, -0.699, -0.699) -- (0.670, -0.679, -0.679);
                \node[below left] at (-0.768, -0.253, -0.253) {$\theta/2$};
        \draw[-stealth, Plum, dashed, opacity=0.3] (Pnext) -- (-0.216, 0.960, 0.000);
        \draw[-stealth, Plum, dashed, opacity=0.3] (Pnext) -- (-0.518, 0.751, 0.000);
        \draw[-stealth, Plum, dashed, opacity=0.3] (Pnext) -- (-0.730, 0.451, 0.000);
        \draw[-stealth, Plum, dashed, opacity=0.3] (Pnext) -- (-0.826, 0.096, 0.000);
        \draw[-stealth, Plum, dashed, opacity=0.3] (Pnext) -- (-0.830, 0.000, 0.000);
        \draw[-stealth, Plum, dashed, opacity=0.3] (Pnext) -- (-0.767, -0.256, -0.256);
        \draw[-stealth, Plum, dashed, opacity=0.3] (Pnext) -- (-0.583, -0.481, -0.481);
        \draw[-stealth, Plum, dashed, opacity=0.3] (Pnext) -- (-0.302, -0.647, -0.647);
        \draw[-stealth, Plum, dashed, opacity=0.3] (Pnext) -- (0.043, -0.736, -0.736);
        \draw[-stealth, Plum, dashed, opacity=0.3] (Pnext) -- (0.411, -0.736, -0.736);

        \draw[vector, NavyBlue] (P) -- (Phat) node[pos=0.65, above left, black]{$v_k$};
        \draw[vector, Plum] (Pnext) -- (Q) node[pos=0.5, right, black]{$v_{k+1}$};

        \fill (P) circle (0.5pt) node[above]{$p_k$};
        \fill (Phat) circle (0.5pt) node[right]{$\hat p_{k+1}$};
        \fill (Pnext) circle (0.5pt) node[below,xshift=-1mm]{$p_{k+1}$};

        \draw[decorate, decoration={brace, amplitude=6pt, raise=9pt}] (Pnext) -- (Phat) node[midway, above=15pt] {$w$};
        \draw[decorate, decoration={brace, amplitude=6pt, raise=9pt}] (Wneg) -- (Pnext) node[midway, above=15pt] {$-w$};

        \node[above] at ($(B)!0.2!(D)$) {$\mathcal T_{k}$};
        \node[right] at (C) {$\mathcal T_{k+1}$};

    \end{tikzpicture}
    };

    \node at (0, -\offy) {(a)};
    \node at (\offx, -\offy) {(b)};
\end{tikzpicture}

%% file: Figures/partransp.tex
\begin{tikzpicture}

    \tikzstyle{triangle} = [ultra thin, gray, rounded corners=0.1pt]
    \tikzstyle{geodesic} = [NavyBlue]
    \tikzstyle{point} = [circle, minimum radius=2pt, black]
    \tikzstyle{vector} = [-stealth, thick]

    \coordinate (V0) at (-2.396906868625721, 1.0481795585910636);
    \coordinate (V1) at (-1.2975402707849137, 0.06910588054655342);
    \coordinate (V2) at (-0.006646638672219152, 1.7013665275540721);
    \coordinate (V3) at (-0.2799599922189223, -1.6825249405893197);
    \coordinate (V4) at (1.4648534216782325, -1.8454682951300705);
    \coordinate (V5) at (1.5170648673289517, -0.03875338944440654);
    \coordinate (V6) at (2.9592631095041253, -0.5635338123497262);
    \draw[triangle, fill=black!10] (V0) -- (V1) -- (V2) -- cycle;
    \draw[triangle, fill=black!32] (V1) -- (V3) -- (V2) -- cycle;
    \draw[triangle, fill=black!16] (V3) -- (V5) -- (V2) -- cycle;
    \draw[triangle, fill=black!10] (V4) -- (V5) -- (V3) -- cycle;
    \draw[triangle, fill=black!18] (V6) -- (V5) -- (V4) -- cycle;

    \coordinate (P0) at (-1.233116321556319, 0.9394963411123829);
    \coordinate (P1) at (-0.6157512229614374, 0.9311888629388901);
    \coordinate (P2) at (-0.12447174792832878, 0.24257435084221546);
    \coordinate (P3) at (1.1368229471533944, -0.38656763787642445);
    \coordinate (P4) at (1.505564637159299, -0.43670516050480734);
    \coordinate (P5) at (1.9787739704239937, -0.8147908959521499);
    \draw[geodesic] (P0) -- (P1) -- (P2) -- (P3) -- (P4) -- (P5);
    \node[above] at (0.5061755996125328, -0.07199664351710455) {$\gamma$};

    \draw[vector, Dandelion] (-1.233116321556319, 0.9394963411123829) -- (-1.305703107365026, 1.7745759130451317) node[above, black] {$n(0)$};

    \draw[vector, Dandelion] (1.9787739704239937, -0.8147908959521499) -- (2.474183906158001, -0.188531673364021) node[above, black] {$n(\ell)$};

    \draw[vector, Green] (-1.233116321556319, 0.9394963411123829) -- (-0.3191152584204783, 0.9271972268608131) node[right, black] {$\dot \gamma(0)$};

    \draw[vector, Green] (1.9787739704239937, -0.8147908959521499) -- (2.707672824907291, -1.397167954246329) node[right, black] {$\dot \gamma(\ell)$};

    \draw[vector, Plum] (-1.233116321556319, 0.9394963411123829) -- (-0.8421824018150167, 0.3750989952630709) node[below, black] {$v$};

    \draw[vector, Plum] (1.9787739704239937, -0.8147908959521499) -- (2.0653161047804804, -1.7097033039018248) node[below, black] {$w$};

    \fill[black] (-1.233116321556319, 0.9394963411123829) circle (1pt) node[left, black] {$p$};

    \fill[black] (1.9787739704239937, -0.8147908959521499) circle (1pt) node[below left, black] {$q$};

\end{tikzpicture}

%% file: algorithms/tv-proj.tex
 \begin{algorithm}[bt]
     \caption{Numerical procedure to perform vector projection on the tangent cone~\eqref{eq:vector-projection}.}
    \label{alg:projection}
 \begin{algorithmic}[1]
    \renewcommand{\algorithmicrequire}{\textbf{Input:}}
    \Require $p \in \mesh$ point defining the tangent cone, $v^\circ \in \R^3$ vector to project

    \renewcommand{\algorithmicensure}{\textbf{Output:}}
    \Ensure $v \in T_p \mesh$

    \State $\mathcal C \gets \emptyset$ empty candidate list

    \For{face $\face_i$ around $p$}
        \State $n \gets $ normal of $\face_i$
        \State $v_\text{pr} \gets (I - n n^\top) v^\circ$ 
        \Comment{plane projection}
        \If{$v_\text{pr} \in T_p\mesh$}
            \Comment{($\dagger$)}
            \State $\mathcal C \gets \mathcal C \cup \{ v_\text{pr} \}$ 
            \Comment{add candidate}
        \EndIf
    \EndFor

    \For{directed edge $e_i$ emitting from $p$} \Comment{($\ddagger$)}
        \State $v_\text{pr} \gets (e_i \cdot v^\circ) e_i$ 
        \Comment{vector projection}
        \If{$v_\text{pr} \cdot v^\circ > 0$}
            \State $\mathcal C \gets \mathcal C \cup \{ v_\text{pr} \}$ 
            \Comment{add candidate}
        \EndIf
    \EndFor

    \State $v \gets \arg \max_{v_\text{pr} \in \mathcal C} v_\text{pr} \cdot v^\circ$
\end{algorithmic}
{
\footnotesize
($\dagger$): generally, a normally-projected vector may extend on the half-plane which does not contain $\face_i$. \\
($\ddagger$): here $e_i$ is a unit norm vector; if $p$ is inside an edge, then there are two directed edges such that $e_1 = -e_2$.
}
\end{algorithm}

%% file: algorithms/gp-threshold.tex
 \begin{algorithm}[bt]
     \caption{Bisection-based algorithm to find the length scale threshold $k_{th}$ for the kernel~\eqref{eq:mesh-kernel} on a mesh $\mesh$.} 
    \label{alg:kernel-bisection}
 \begin{algorithmic}[1]
    \renewcommand{\algorithmicrequire}{\textbf{Input:}}
    \Require $\{p_i \in \mesh\}_{i=1}^m$ test points, $\kappa_0$ initial length scale value

    \renewcommand{\algorithmicrequire}{\textbf{Parameters:}}
    \Require $\delta$: numerical accuracy

    \renewcommand{\algorithmicensure}{\textbf{Output:}}
    \Ensure $\kappa_{th}$
    \State initialise the distance $(d_{ij}) = D \in \R^{m\times m}$
    \State $d_{ij} \gets d_g(p_i, p_j) \quad \forall i, j = 1, \dots, p$
    \State $\kappa_{min} \gets \kappa_0$, $\ \kappa_{max} \gets 2\kappa_0$

    \While {$\exp(-D / 2\kappa_{min}^2) \succ 0$}
    \Comment ($\dagger$)
        \State $\kappa_{min} \gets 2\kappa_{min}$, $\ \kappa_{max} \gets 2\kappa_{max}$
    \EndWhile

    \While {$\exp(-D / 2\kappa_{min}^2) \prec 0$}
    \Comment ($\dagger$)
        \State $\kappa_{min} \gets \kappa_{min} / 2$, $\ \kappa_{max} \gets \kappa_{max} / 2$
    \EndWhile

    \While {$|\kappa_{max} - \kappa_{min}| > \delta$}
        \State $\kappa_{mid} \gets (\kappa_{max} + \kappa_{min}) / 2$
        \If {$\exp(-D / 2\kappa_{mid}^2) \succ 0$}
        \Comment ($\dagger$)
            \State $\kappa_{min} \gets \kappa_{mid}$
        \Else
            \State $\kappa_{max} \gets \kappa_{mid}$
        \EndIf
    \EndWhile
    \State $\kappa_{th} \gets \kappa_{min}$
\end{algorithmic}
{
\footnotesize
($\dagger$): matrix obtained by evaluating component-wise the exponential function
}
\end{algorithm}

%% file: algorithms/rfm.tex
 \begin{algorithm}[bt]
    \caption{Riemannian Flow Matching on meshes; training procedure.}
    \label{alg:rfm}
 \begin{algorithmic}[1]
    \renewcommand{\algorithmicrequire}{\textbf{Input:}}
    \Require $\pi_\textrm{ref} \in \mathcal P$ reference distribution, $\pi_\textrm{data} \in \mathcal P$ data distribution, $\eta$ scheduler

    \renewcommand{\algorithmicensure}{\textbf{Output:}}
    \Ensure $v_t(\cdot, \Theta) \in T_{(\cdot)}\mesh$

    \State Initialise parameters $\Theta$ of $v_t$

    \While {not converged}
        \State sample time $t \backsim \mathcal U(0, 1)$
        \State sample training example $p_1 \backsim \pi_\textrm{data}$
        \State sample noise $p_0 \backsim \pi_\textrm{ref}$
        \State construct geodesic $\gamma \in [0, 1] \rightarrow \mesh$  from $p_0$ to $p_1$
        \State $p_t \gets \gamma(t)$
        \State $\dot \gamma_t \gets \dot \gamma(t)$
        \State $u_t \gets \ell\, \dot\gamma_t / |\dot\gamma_t| $ 
        \State $\Theta = \texttt{optimiser\_step}\big( | v_t(p_t, \Theta) - u_t |^2 \big)$
    \EndWhile
\end{algorithmic}
\end{algorithm}

%% file: algorithms/rfm-sampling.tex
 \begin{algorithm}[bt]
    \caption{Riemannian Flow Matching on meshes; generating samples from a trained NN.}
    \label{alg:rfm-sampling}
 \begin{algorithmic}[1]
    \renewcommand{\algorithmicrequire}{\textbf{Input:}}
    \Require $v_t(p, \Theta)$ learned conditional vector field, $\mesh$ mesh, $\delta_t$ integration step size

    \renewcommand{\algorithmicensure}{\textbf{Output:}}
    \Ensure $q \backsim \pi_\textrm{data}$

    \State $t \gets 0$
    \State $p \gets $ sample from $\pi_\textrm{ref}$ on $\mesh$

    \While {$t < 1$}
        \State $v^\circ \gets v_t(p, \Theta)$
        \State $v \gets \textrm{Proj}_p(v^\circ)$
        \State $p \gets \Exp_p(\delta_t \, v)$
        \State $t \gets t + \delta_t$
    \EndWhile
    \State $q \gets p$
\end{algorithmic}
\end{algorithm}

%% file: Sections/simulation.tex
\section{Simulation Results}
\label{sec:simulations}

In this section, we evaluate and compare the proposed methodology against state-of-the-art approaches.
Section~\ref{sec:meshdmp-comparison} compares our DMP with the MeshDMP framework~\cite{DalleVedove25}, Sec~\ref{sec:gp-comparison} compares our geodesic-based squared exponential kernel with the one computed from the Laplace-Beltrami operator introduced in~\cite{Borovitskiy2020}, while Sec.~\ref{sec:fm-comparison} compares our Mesh Flow Matching against the RFM approach in~\cite{Chen2023}.

Our method has been implemented fully in C++ and it relies on the CGAL library for polyhedral mesh operations~\cite{cgal-library, cgal-shortest-path}. 
Experiments in Sec.~\ref{sec:meshdmp-comparison} and~\ref{sec:gp-comparison} have been carried out on a commercial laptop equipped with an \textit{Intel Core i7 Ultra 155H}, 24GB of RAM, and a \textit{NVidia RTX 4060} graphics card with 8GB of VRAM. For the experiments on Flow Matching (Sec.~\ref{sec:fm-comparison}), we used a workstation equipped with an \textit{AMD Ryzen 9900X} CPU with with $12$ cores, $96$GB of RAM, and a \textit{NVidia RTX 5090} graphics card with $32$GB of VRAM.

For testing, we used several meshes, whose main properties are summarised in~\autoref{tab:meshes}.

\input{Tables/meshes.tex}

\subsection{Comparison against MeshDMP}
\label{sec:meshdmp-comparison}

We compare against the MeshDMP~\cite{DalleVedove25} framework on the representative task of learning an 8-shaped trajectory demonstrated in the trivial, flat surface, and generalise the policy to the Stanford Bunny mesh.
Both approaches utilise same DMP parameters ($\alpha = 48$, $\beta = \alpha/4$, $\gamma=5$, and $20$ basis) and are initialised at the same initial state, allowing the direct behaviour comparison between MeshDMP and the rhythmic DMP~\eqref{eq:rhythdmp}.

\begin{figure}[bt]
    \centering
    \includegraphics[width=\linewidth]{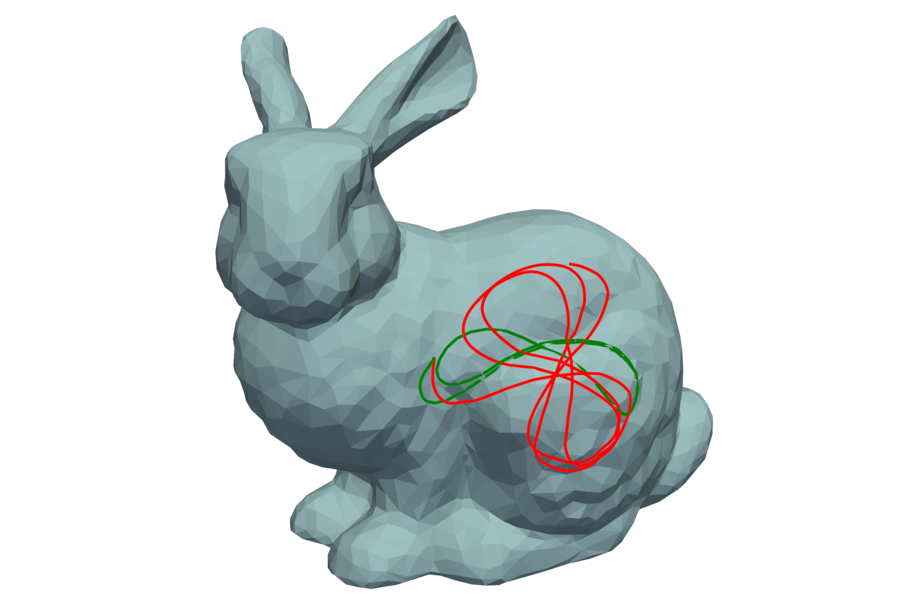}
    \caption{
        DMP transferability comparison against~\cite{DalleVedove25}. 
        Both DMP methods have been trained on the same 8-shaped demonstration on the flat plane, and have been generalised onto the Stanford bunny mesh, with equal initial state and centre.
        The green line represent our method, while the red one is the baseline MeshDMP result.
    }
    \label{fig:meshdmp-forcing-comparison}
\end{figure}

As depicted in Fig.~\ref{fig:meshdmp-forcing-comparison}, both methods quickly converge to the limit cycle induced by the 8-shaped trajectory in the initial phase of execution.
Differences, however, arise when we let system integrate their dynamics over a longer time horizon.
Particularly, we observe that DMP~\eqref{eq:rhythdmp} remains faithful to the limit cycle, while MeshDMP exhibits an undesired rotational behavior around the fixed motions centre.
Such a difference can be attributed on how the two methods encode the forcing term $f$ used during DMP integration. 
MeshDMP encodes this term using a local parameterisation that fails in capturing the dynamics change induced by the mesh curvature. 
Our method, instead, overcomes this limitation by encoding $f$ in a fixed coordinate system and use a geometrically-consistent parallel transport operation to ensure that the forcing term is correctly applied at the actual DMP position state.

\begin{figure}[bt]
    \centering
    \includegraphics[width=\linewidth]{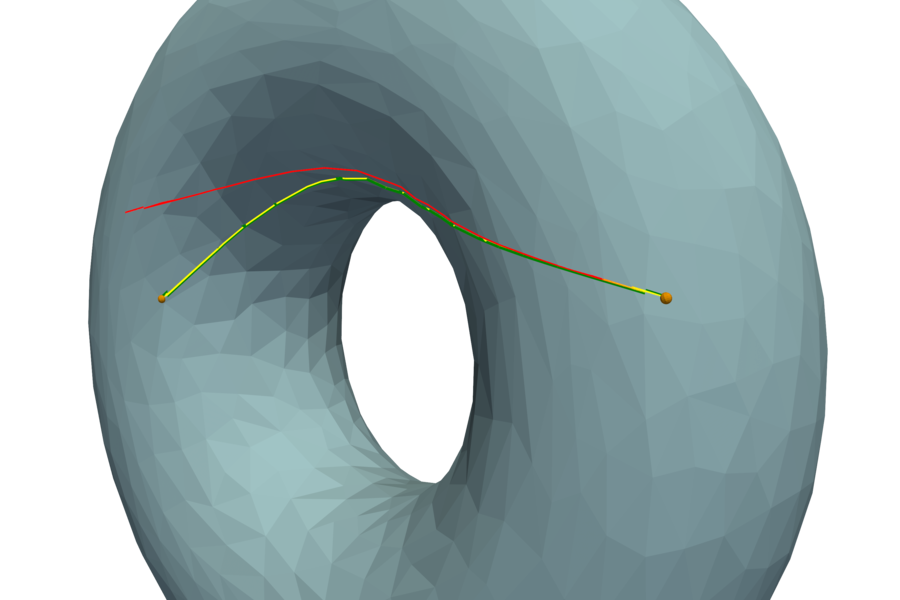}
    \caption{Exponential map comparison against~\cite{DalleVedove25}. 
    First we compute the geodesic path (green) between the two orange points, and the logarithmic map $v$ between them.
    Then we compute the exponential map of $v$ with Algorithm~\ref{alg:exp-map} (yellow) and the MeshDMP-proposed algorithm (red).
    The reconstructed path by our method coincides, up to numerical precision, with the reference geodesic.
    }
    \label{fig:meshdmp-expmap-comparison}
\end{figure}

Additionally, other known limitations of MeshDMP may contribute to these behaviours. 
Remarkably, Dalle Vedove et al. relied on a local approximation of the exponential map based on surface normal projection, which is only accurate for sufficiently small vectors and restricts its applicability to surfaces whose relative angle between edge-adjacent faces is less then $90^\circ$.
The local approximation of MeshDMPs exponential map is shown in Fig.~\ref{fig:meshdmp-expmap-comparison}, where we compare the two integration schemes applied to a long vector, showing how exponential map~\eqref{eq:expmap} can faithfully reconstruct a geodesic on the Mesh.

\subsection{Gaussian Processes}
\label{sec:gp-comparison}

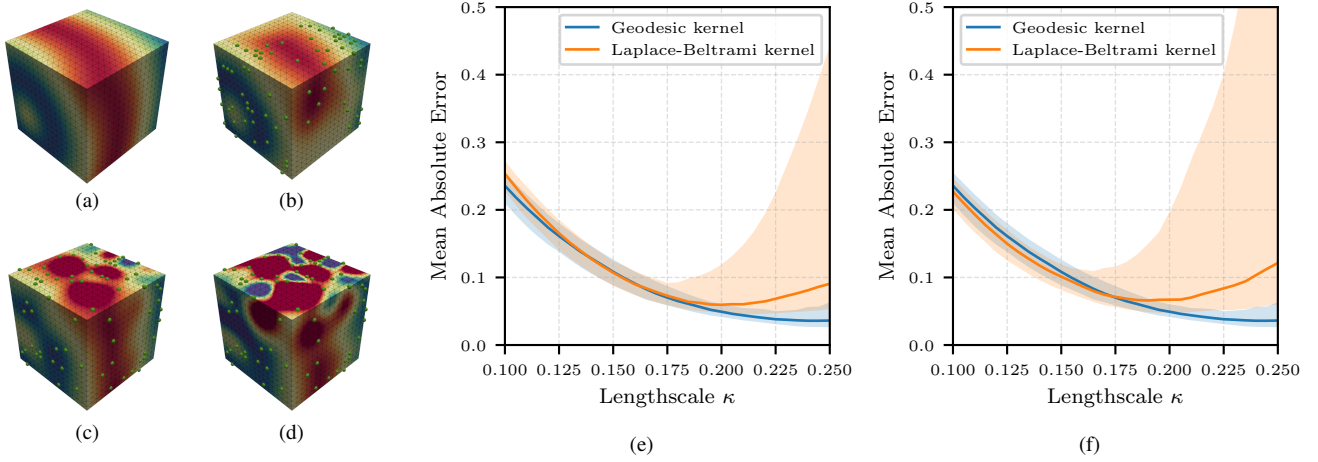
\begin{figure*}[bt] 
    \centering
    \input{Figures/cube-gp.tex}
    \caption{ Experiment on a mesh composed of three vertex-adjacent
      faces of a cube.  The ground truth (a) is evaluated at the
      vertices of a mesh upsampled to $817$ vertices and $1\,536$
      faces by computing the sine of the geodesic distance from a
      fixed point.  The noise-free GP predictions are obtained using
      70 ground-truth samples with (b) the geodesic based
      kernel~\eqref{eq:mesh-kernel} and (c--d) the Laplace-Beltrami
      kernel~\eqref{eq:lb-kernel} with $\kappa=0.18, \sigma=1$.
      Panels (e--f) report the median MAE together with the 10th-90th
      percentile range as a function of the length scale, computed over
      100 experiments, each using 70 randomly selected test vertices.
      Figures (c) and (e) use the upsampled mesh, while (d) and (f)
      use a further upsampled mesh with $12\,481$ vertices and
      $24\,576$ faces.  (b--d) show the worst-performing samples in
      their respective category for the selected length scale.  }
    \label{fig:cube-comparison}
\end{figure*}

In this section, we evaluate the performance of the proposed geodesic-based kernel for GPR on discrete surfaces. 
We compare our approach against the state-of-the-art work by Borovitskiy et al.~\cite{Borovitskiy2020}, where authors used the spectral properties of the Laplace-Beltrami operator to define kernels on compact manifolds, including the discrete space of meshes. 
Starting by the consideration that, in Euclidean space, GPs are the solution of a stochastic differential equations that depends on the Laplacian $\Delta$~\cite{Whittle63}, they generalise such formulation to Riemman manifolds by replacing the Laplacian with the Laplace-Beltrami operator $\Delta_g$;
by computing the orthonormal basis given by the eigenfunction-eigenvalue pairs $\{(f_n, \lambda_n)\}$ such that
\begin{equation} \label{eq:lb-decomposition}
    -\nabla_g f = \sum_{n=0}^\infty \lambda_n \langle f, f_n \rangle f_n,
\end{equation}
authors define the squared exponential kernel to be
\begin{equation} \label{eq:lb-kernel}
    k_\Delta(x, x') = \frac{\sigma^2}{C} \sum_{n=0}^\infty e^{-\frac{\kappa^2}{2} \lambda_n} f_n(x) f_n(x').
\end{equation}
Here, $\kappa$ and $\sigma$ are respectively the length scale and variance hyperparameters as in~\eqref{eq:mesh-kernel}, and $C$ is a normalising constant; for computational purposes, the evaluation of the kernel~\eqref{eq:lb-kernel} is truncated.

While the kernel~\eqref{eq:lb-kernel} offers the advantage of being
positive-definite for any value of $\kappa$, it suffers from at least
two major limitations: a) it implicitly assumes the mesh to be
sufficiently smooth, effectively considering it as an approximation of
a surface; b) the Laplace-Beltrami operator $\Delta_g$ can only be
evaluated at the vertices of such a mesh.  It follows that predicting
function values at arbitrary points, specifically ones located at the
interior of the faces, either requires additional interpolation steps,
thus introducing approximation errors, or upsampling the mesh to let
the test point become a new vertex for the mesh, effectively
increasing the model complexity.  The geodesic-based
kernel~\eqref{eq:mesh-kernel}, instead, automatically solves this
problem by enabling the computation of geodesics at arbitrary points.

As an illustrative example, we consider the problem of performing GPR
on a non-smooth, trivial surface composed by 3 vertex-adjacent faces
of a cube, shown in Fig.~\ref{fig:cube-comparison}.  A minimal
representation of this mesh consists of 7 vertices and 3
faces\footnote{For the experiments, we used a triangulated version of
  the mesh which actually has 6 faces.}; however, to enable a valid
application of the Laplace-Beltrami-based kernel~\eqref{eq:lb-kernel},
we rely on midpoint subdivision to upsample the mesh up to $817$
vertices and $1\,536$ faces.  We additionally test GPR on a further
upsampled mesh with $12\,481$ vertices and $24\,576$ faces.  To
evaluate $k_\Delta$, we used the \textsc{GeometricKernels}
package~\cite{mostowsky2024}, and use the first $1\,000$ eigenfunction
as basis.  By constructing a ground-truth function on the mesh, we
compute a noise-free GP prediction~\eqref{eq:gp-prediction} using both
kernels, and evaluate regression performance using the mean absolute
error (MAE) metric.  We run 100 experiments, each randomly selecting
70 vertices from the upsampled mesh as training data, enabling a
statistical characterisation of performance across different
length scale values.  While the two methods exhibit comparable
performance at small length scales, we observe that the
Laplace-Beltrami-based kernel becomes unstable for larger values of
$\kappa$, with the MAE in some cases degrading to unacceptable levels.
In contrast, the geodesic-based kernel remains well behaved, provided
that the critical length $\kappa_{th}$ is not reached.  Moreover, as
shown in Fig~\ref{fig:cube-comparison}(d), further increasing the
polygonal density of the mesh, which is used to discretise the
Laplace-Beltrami operator on a non-smooth surface, leads to degraded
performance.  This behaviour is a consequence of the assumption of a
smooth underlying manifold implicitly made by the Laplace-Beltrami
operator: increasing the mesh resolution in the presence of geometric
discontinuities, amplifies the discretisation errors, ultimately
resulting in a breakdown of the GPR model.

\begin{figure}[bt] 
    \centering
    \input{Figures/dragon-gp.tex}
    \caption{
        Experiment on the Stanford Dragon mesh, downsampled to $9\,983$ faces and $4\,898$ vertices. (a) Kernel with length scale $\kappa = 0.018$ evaluated at some test points; (b) ground truth computed as the sine of the geodesic distance from a fixed point; (c) noise-free GPR with $\kappa = 0.02$ using the geodesic based kernel~\eqref{eq:mesh-kernel}; (d) median and 10th-90th percentile range MAE across 100 experiments sampling each $48$ points randomly on the mesh.
    }
    \label{fig:dragon-gp}
\end{figure}
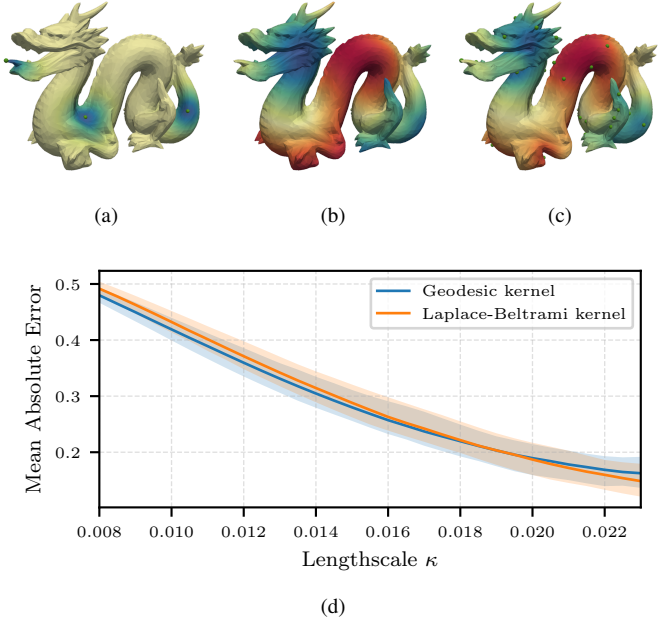

A similar experiment, Fig.~\ref{fig:dragon-gp}, has been ran on the
Stanford Dragon mesh with $9\, 983$ faces and $4\,898$ vertices.  Also
in this case, our proposed kernel performs slightly better than the
Laplace-Beltrami based solution in the whole spectrum of length scales
below the critical value of $k_{th} = 0.023$, which is assessed
through the Algorithm~\ref{alg:kernel-bisection}.

\paragraph*{Computation performance} 
To fully compare the methods, we also evaluate their computational
performance both in terms of time required to construct the necessary
data structures and to compute the covariance matrices. The
summarising results are reported in~\autoref{tab:gp-times}.

Within our method, it is necessary to construct a data structure for
each set of input point for the kernel; upon their construction, the
evaluation time of the covariance matrices is approximately linear
w.r.t. the number of test points.  On the other hand, Laplace-Beltrami
based method require the one-time eigenfunction decomposition---which
is time demanding---but then provides an overall faster inference.

Our method shows its maximum strength in all cases in which the mesh
is not known upfront, e.g., for time-varying surfaces, it is small to
medium sized (up to $5\,000$ faces), and it is limited to the
computation of the shortest path on $\mesh$, which is delegated to
external libraries.

\input{Tables/gp-times.tex}

\subsection{Mesh Flow Matching}
\label{sec:fm-comparison}
\begin{figure}
    \centering
    \begin{tikzpicture}[every node/.style={font=\footnotesize}]
        \node[align=center, anchor=east] (ref) { \includegraphics[height=4cm]{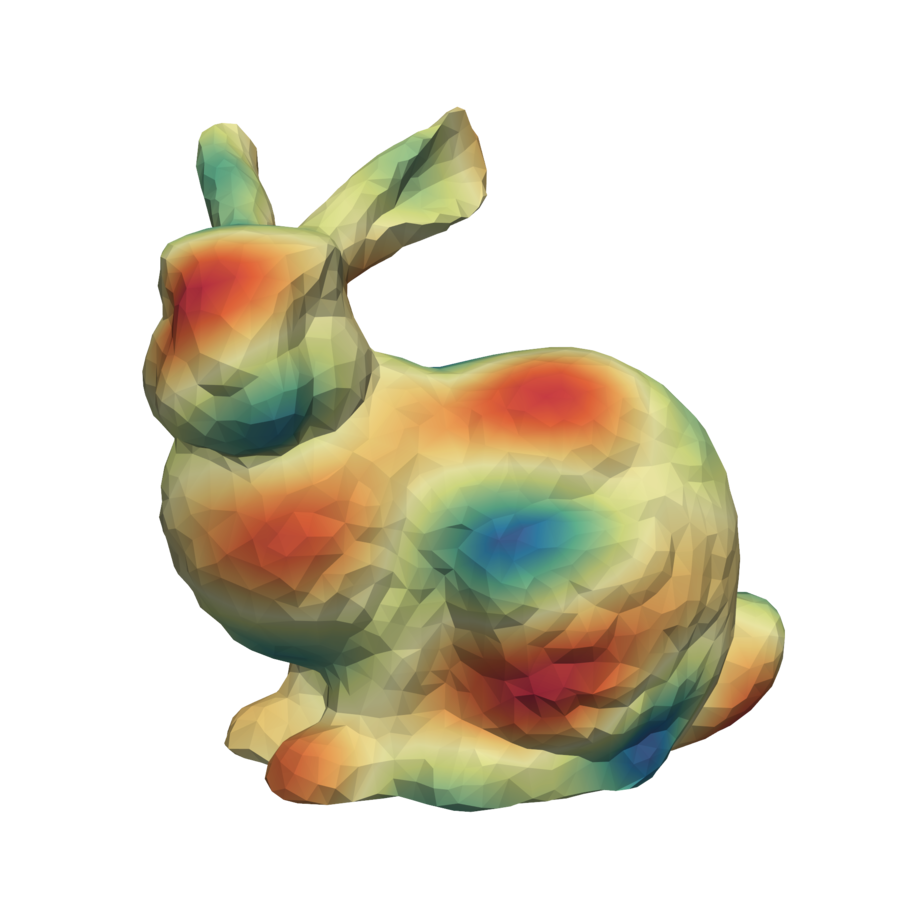} \\ (a) };
        \node[align=center, anchor=west] (x1s) { \includegraphics[height=4cm]{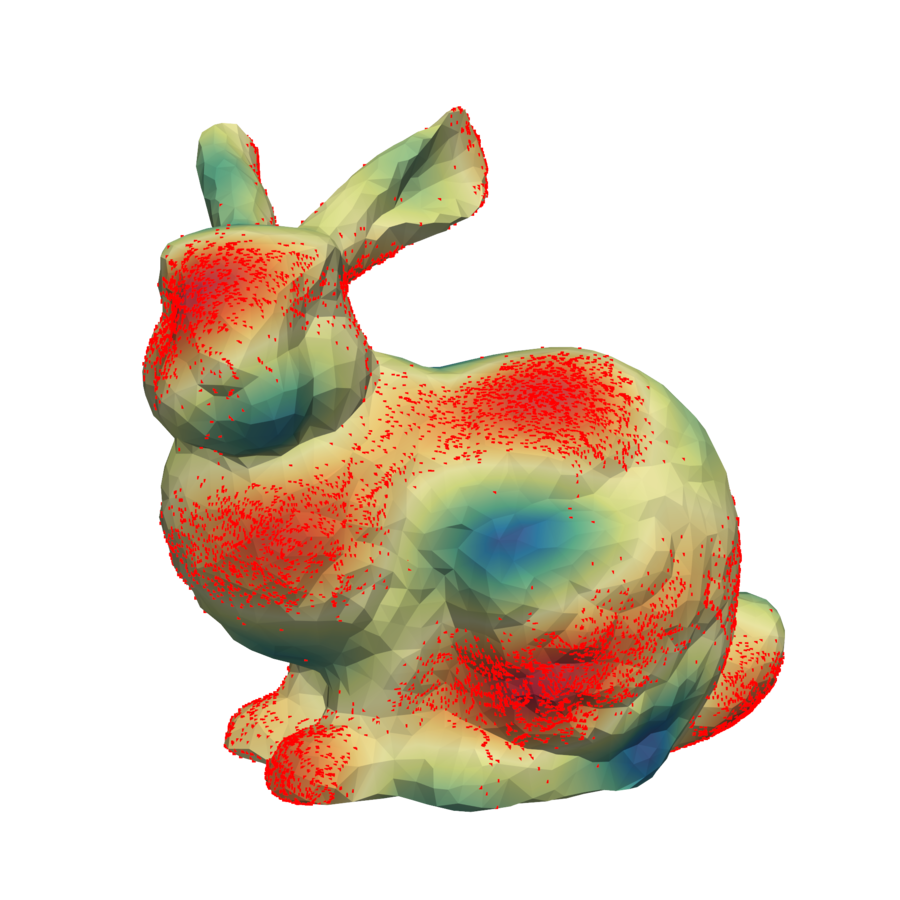} \\ (b) };
    \end{tikzpicture}
    \caption{Target distribution (a) based on the 50th Laplace-Beltrami eigenfunction and $20\,000$ samples (b) generated with our geodesic flow-matching on the Stanford Bunny.}
    \label{fig:flow-matching}
\end{figure}

To fairly compare our proposed Mesh Flow Matching against the baseline, we integrated our mesh operators in the official implementation of the RFM framework provided by~\cite{Chen2023}.
As in the reference paper, the surface used for the experiment is the Stanford Bunny, decimated to $5\,000$ faces, and Spot the Cow.
To construct the non-trivial target distribution that must be learned from the generative model, $500\,000$ samples are drawn from the mesh by computing the $i$-th eigenfunction of the Laplace-Beltrami operator~\eqref{eq:lb-decomposition} and sampling points on $\mesh$ proportionally to the thresholded eigenfunction value. 
The dataset is split into training, validation, and test sets using an 80--10--10 ratio.
An example of target data distribution and sample reconstruction with our mesh RFM method is reported in Fig.~\ref{fig:flow-matching}.

To evaluate the regression performance, we compute the negative log-likelihood (NLL) by averaging the log density $\log \pi_\textrm{data}(p)$ of all samples in the test dataset.
Following~\cite{Chen2018}, log-density is computed as
\begin{equation} \label{eq:nll}
    \log \pi_\textrm{data}(p) = \log \pi_\textrm{ref} (p_0) - f_0(p),
\end{equation}
where $f_t$ is obtained solving backward in time from $t = 1$ to $t=0$ the ODE
\begin{equation} \label{eq:nll-ode}
    \frac{\d}{\d t} \begin{pmatrix} 
        p_t \\ f_t(p) 
    \end{pmatrix} = \begin{pmatrix} 
        v_t(p_t) \\ - \textrm{div}_{\mathfrak g}(v_t)(p_t)
    \end{pmatrix} 
\end{equation}
subject to the initial condition $\big(p_1, f_1(p)\big)^\top = (p, 0)^\top$; here, $\textrm{div}_{\mathfrak g}$ is the Riemannian divergence.

\input{Tables/rfm-nll.tex}

Table~\ref{tab:rfm-nll} reports the test NLL---evaluated at the model checkpoint with lowest validation NLL---of our method, and the RFM baseline which uses diffusion and biharmonic premetrics computed on the first $200$ eigenfunctions.
All models have been trained with the same parameters and seeds, using a Dense NN with $6$ layers of $512$ neurons each with the Swish activation function~\cite{Ramachandran18}, and using the ADAM optimiser with a learning rate of $10^{-4}$.
Across all experiments, our method consistently achieves lower test NLL, indicating improved generative performance when using geometry-aware operators defined directly on the mesh.

\paragraph*{Computation performance} 
The claimed advantage of Laplace-Beltrami based spectral operators, is
that they limit the computation time by only considering a fixed,
relatively small number of eigenfunctions---200 in this case.  However,
since the integration of the spectral measures doesn't guarantee the
path to lie on the mesh, closest-point projections are required by the
baseline algorithm at every step, effectively slowing down the
integration process.  On the other hand, our method---by using the
exponential map~\eqref{eq:expmap}---is always guaranteed, during ODE
integration, to generate paths which are consistent with the
underlying surface.  Additionally, during the training
(Algorithm~\ref{alg:rfm}), the time-intensive geodesic path
computation between samples $p_0$ and $p_1$ is amortised by the cheap
upsampling over time of the vector field $u_t$.

From our experience, performing $100\,000$ update steps of the network with batches of $128$ samples, and calling a validation routine every $2\,000$ steps to save model checkpoints, typically takes 1 day and 6 hours for the baseline method, while the proposed method takes on average $20$ hours.

%% file: Tables/meshes.tex
\begin{table}
    \centering
    \caption{
        Main meshes used for experiments and corresponding number of faces and vertices.
        Upsampling achieved by midpoint subdivision.
    }
    \label{tab:meshes}
    \begin{tabular}[t]{c c c}
        \toprule
        Mesh & n. of faces & n. of vertices \\
        \midrule
        3 faces of a cube & $6$ & $7$ \\
        3 faces of a cube ($4\times$ upsampling) & $1\,536$ & $817$ \\
        3 faces of a cube ($6\times$ upsampling) & $24\,576$ & $12\,481$ \\
        Spot the Cow~\cite{Crane2013} & $5\,856$ & $2\,930$ \\
        Stanford Dragon~\cite{StanfordDragon} & $9\,983$ & $4\,898$ \\
        Stanford Bunny~\cite{StanfordDragon} & $5\, 000$ & $2\,502$ \\
        \bottomrule
    \end{tabular}
\end{table}

%% file: Figures/cube-gp.tex
\begin{tikzpicture}[every node/.style={font=\footnotesize}]

    \def\figh{25mm}
    \def\offy{18mm}

    \node (samples) {
    \begin{tikzpicture}
        \node[align=center, anchor=south east] { \includegraphics[height=\figh]{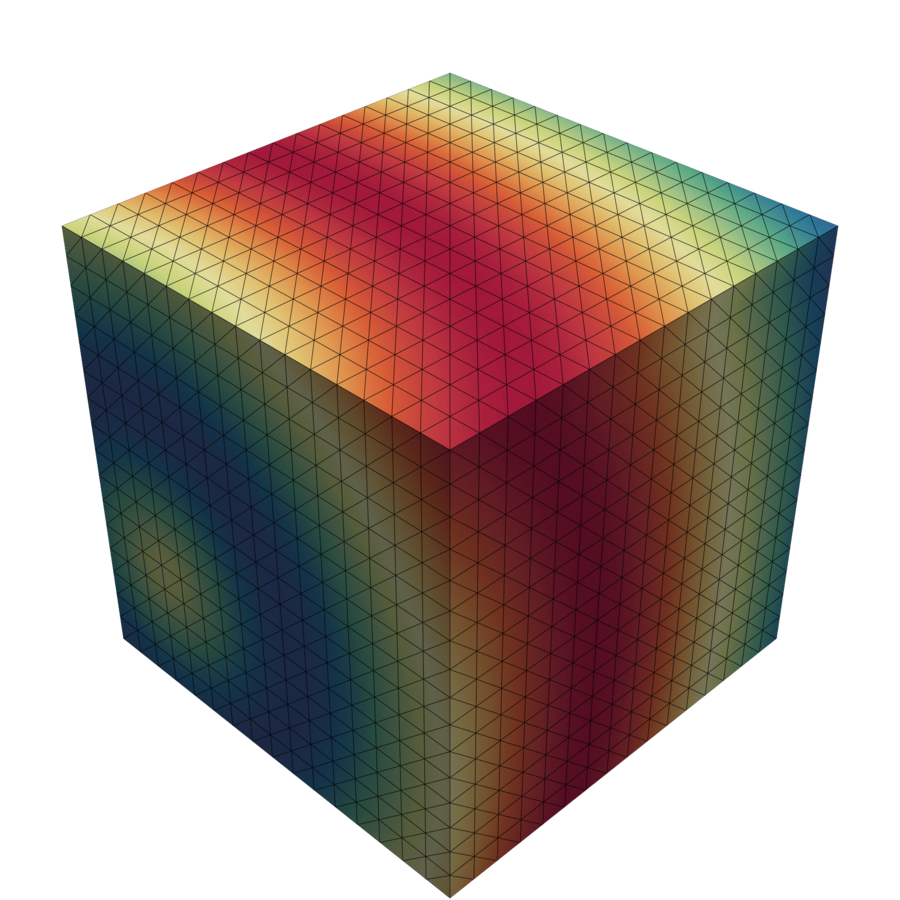} \\ (a)};
        \node[align=center, anchor=south west] { \includegraphics[height=\figh]{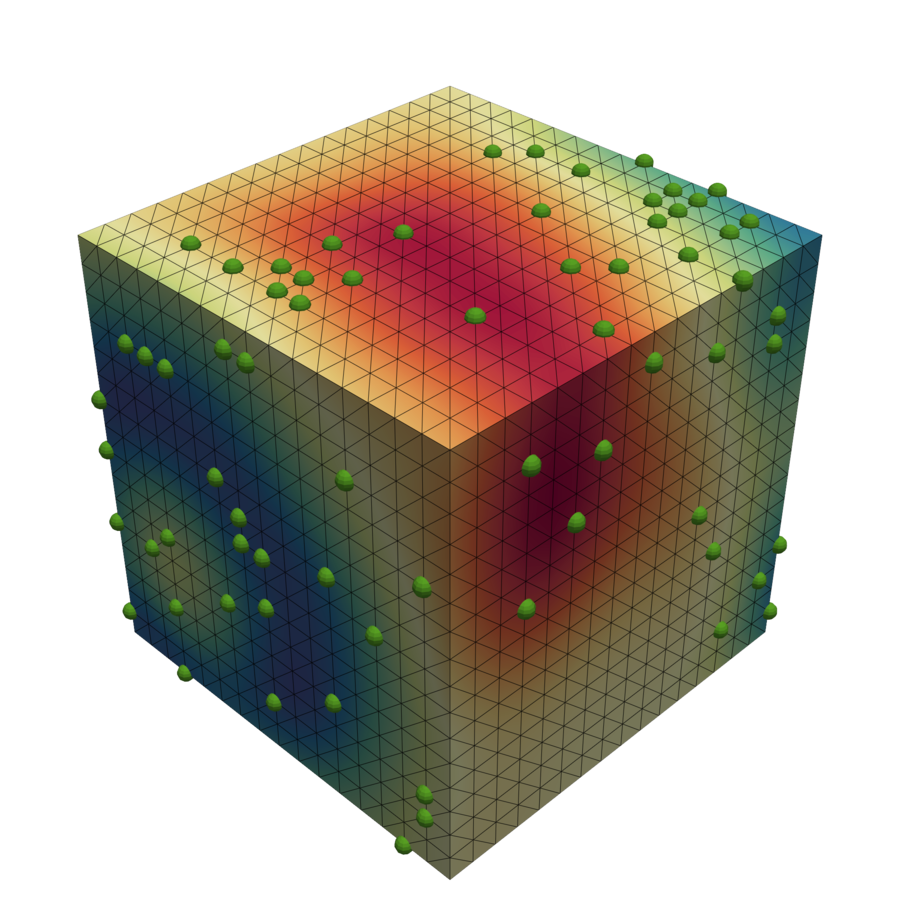} \\ (b)};
        \node[align=center, anchor=north east] { \includegraphics[height=\figh]{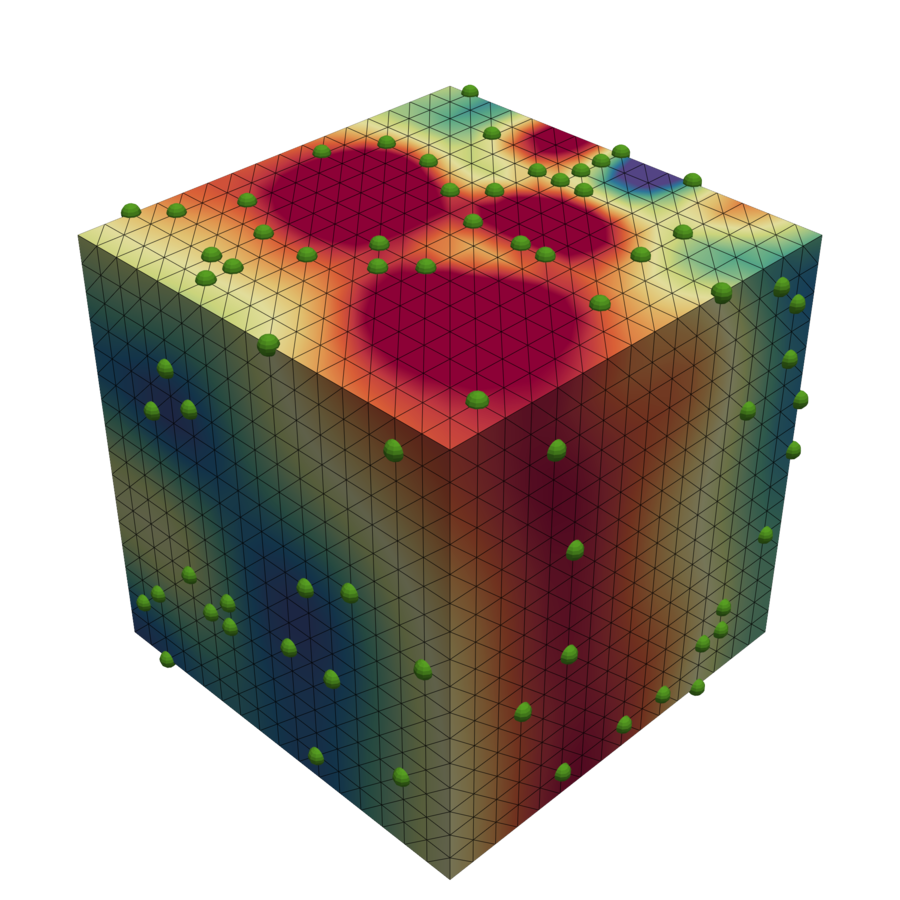} \\ (c)};
        \node[align=center, anchor=north west] { \includegraphics[height=\figh]{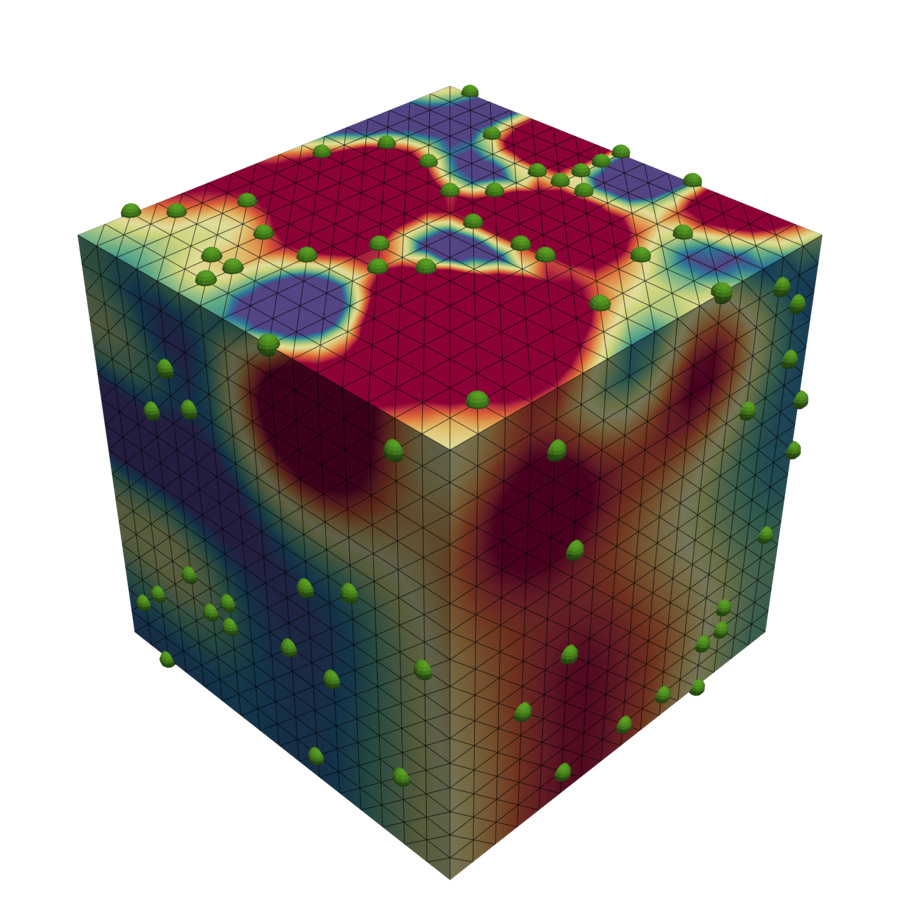} \\ (d)};
    \end{tikzpicture}
    };

    \node[align=center, anchor=west] (maeBase) at (samples.east) { \includegraphics{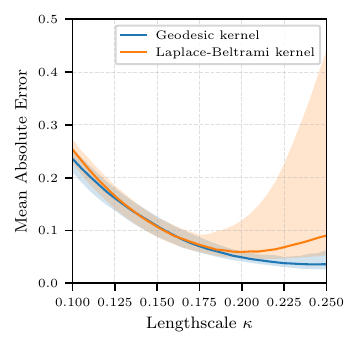} \\ (e)};
    \node[align=center, anchor=west] (maeUpsampeld) at (maeBase.east)[xshift=-4mm] { \includegraphics{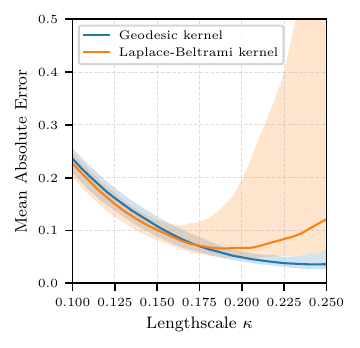} \\ (f)};
\end{tikzpicture}

%% file: Figures/dragon-gp.tex
\begin{tikzpicture}[every node/.style={font=\footnotesize}]

    \def\figh{28mm}
    \def\offy{18mm}

    \node (samples) {
    \begin{tikzpicture}
        \node[align=center] (gt) { \includegraphics[height=\figh]{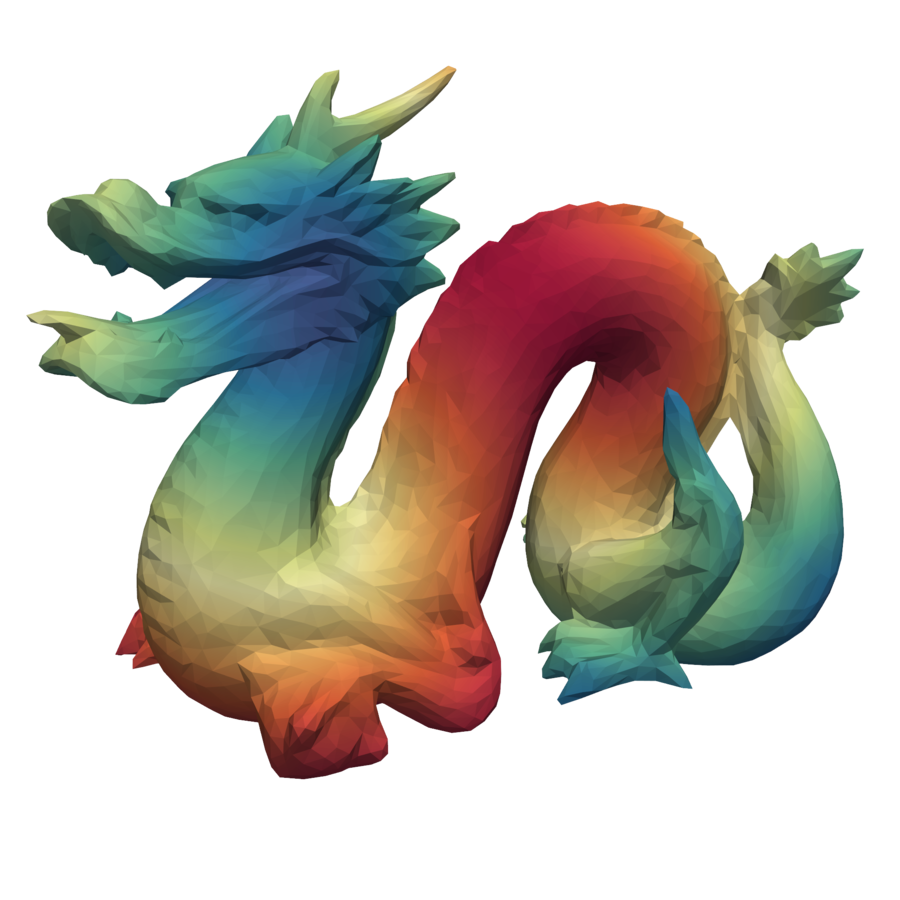} \\ (b)};
        \node[align=center, anchor=east] at (gt.west) { \includegraphics[height=\figh]{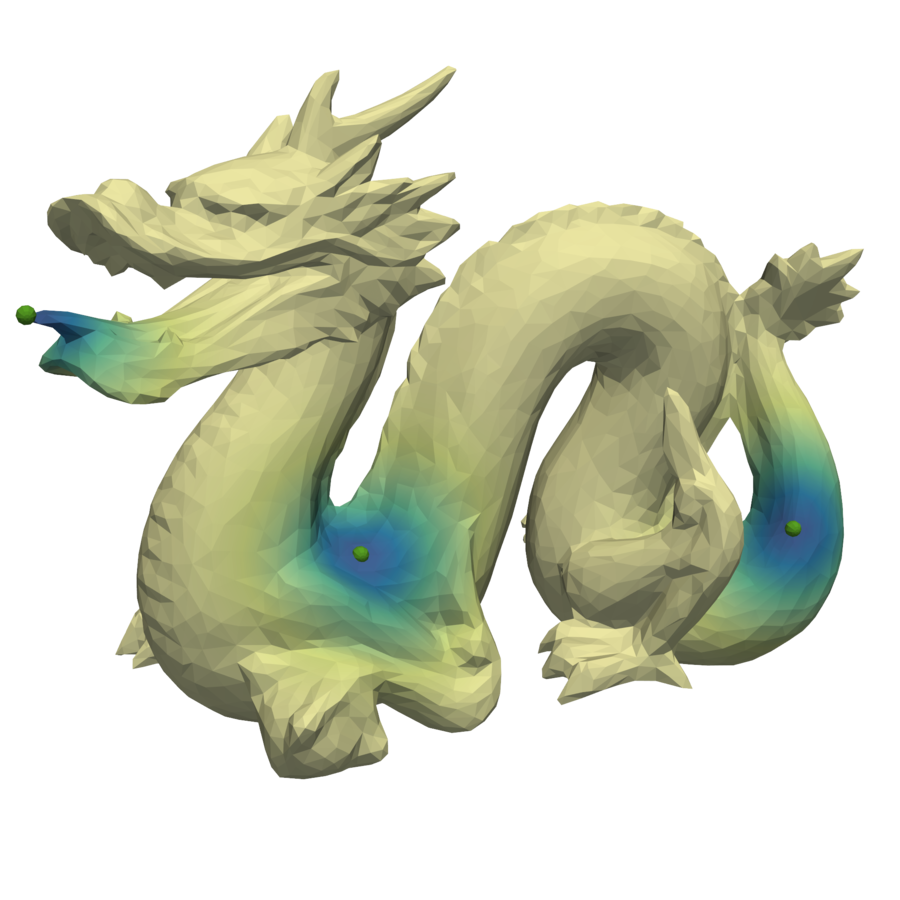} \\ (a)};
        \node[align=center, anchor=west] at (gt.east) { \includegraphics[height=\figh]{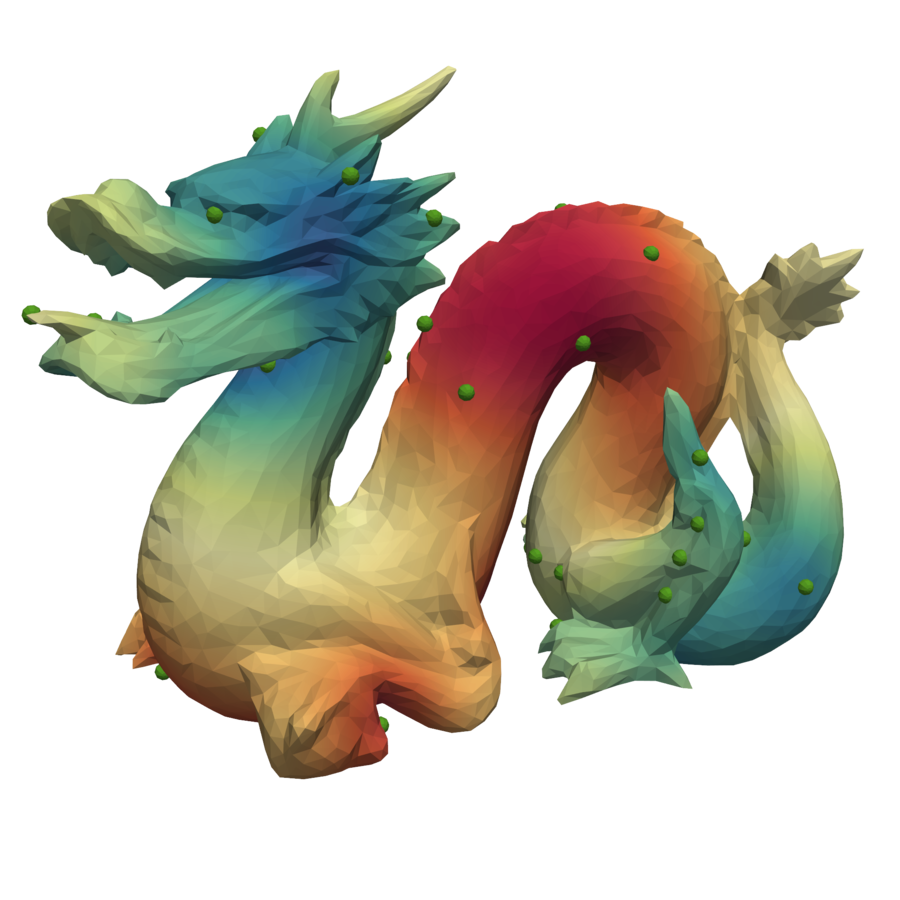} \\ (c)};
    \end{tikzpicture}
    };

    \node[align=center, anchor=north] (maeBase) at (samples.south) { \includegraphics{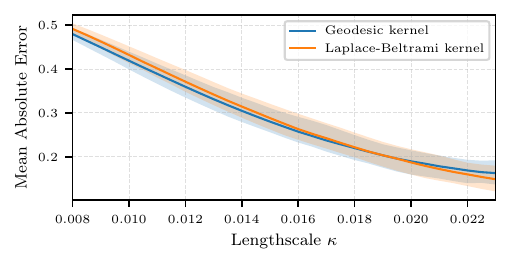} \\ (d)};
\end{tikzpicture}

%% file: Tables/gp-times.tex
\begin{table}
    \centering
    \caption{
        Median computation time of covariance matrices for based on the number of test points $m$ on some meshes.
        For each mesh, we report the eigenfunction computation time; 
        $(\cdot)$ here represent the set of all vertices of the mesh.
    }
    \label{tab:gp-times}
    \begin{tabular}[t]{c c c c c c}
        \toprule
        & \multicolumn{3}{c}{Geodesic Kernel} & \multicolumn{2}{c}{Laplace-Beltrami Kernel} \\
        $m$ & setup & $K_{PP}$ & $K_{(\cdot)P}$  & $K_{PP}$ & $K_{(\cdot)P}$ \\
        \toprule
        \multicolumn{6}{c}{Cube ($240\,$ms and $87\,$s)} \\
        \midrule
        $20$ & $0.343\,$ms & $0.234\,$ms & $4.9\,$ms & $0.739\,$ms & $4.3\,$ms \\
        $40$ & $0.566\,$ms & $0.554\,$ms & $9.08\,$ms & $1.07\,$ms & $5.24\,$ms \\
        $80$ & $0.695\,$ms & $1.13\,$ms & $9.88\,$ms & $1.37\,$ms & $5.59\,$ms \\
        $160$ & $1.37\,$ms & $4.23\,$ms & $19.2\,$ms & $2.82\,$ms & $7.49\,$ms \\
        $320$ & $3.39\,$ms & $14.1\,$ms & $32.2\,$ms & $6.14\,$ms & $11.2\,$ms \\
        \midrule
        \multicolumn{6}{c}{Spot the Cow ($27\,$s)} \\
        \midrule
        $20$ & $472\,$ms & $0.694\,$ms & $15.1\,$ms & $0.868\,$ms & $9.18\,$ms \\
        $40$ & $921\,$ms & $0.941\,$ms & $29.6\,$ms & $0.967\,$ms & $9.98\,$ms \\
        $80$ & $1.73\,$s & $3.45\,$ms & $59\,$ms & $1.21\,$ms & $11\,$ms \\
        $160$ & $3.43\,$s & $11.9\,$ms & $112\,$ms & $2.14\,$ms & $18.6\,$ms \\
        $320$ & $6.74\,$s & $40.1\,$ms & $224\,$ms & $4.44\,$ms & $29.1\,$ms \\
        \midrule
        \multicolumn{6}{c}{Stanford Dragon ($49\,$s)} \\
        \midrule
        $20$ & $565\,$ms & $0.75\,$ms & $37.4\,$ms & $0.863\,$ms & $17.8\,$ms \\
        $40$ & $1.05\,$s & $1.52\,$ms & $67.9\,$ms & $0.925\,$ms & $19.4\,$ms \\
        $80$ & $2.15\,$s & $5.14\,$ms & $128\,$ms & $1.35\,$ms & $23.3\,$ms \\
        $160$ & $4.22\,$s & $19.3\,$ms & $242\,$ms & $2.09\,$ms & $42.9\,$ms \\
        $320$ & $8.37\,$s & $61.8\,$ms & $477\,$ms & $4.43\,$ms & $147\,$ms \\
        \bottomrule
    \end{tabular} \vspace{1mm}

    \footnotesize
    For the 3 vertex-adjacent faces of the cube, we report the eigenfunction computation time both for the $4\times$ and $6\times$ upsampling.
\end{table}

%% file: Tables/rfm-nll.tex
\begin{table}
    \centering
    \caption{
        Test NLL on the Stanford Bunny with dataset generated from the $i$-th eigenfunction. Lower is better.
    }
    \label{tab:rfm-nll}
    \begin{tabular}[t]{c c c c}
        \toprule
        & Geodesic (ours) & Diffusion & Biharmonic \\
        \toprule 
        \multicolumn{4}{c}{Stanford Bunny} \\
        \midrule
        $i = 10$ & $0.81\pm 0.1$ & $1.16 \pm 0.02$ & $1.06 \pm 0.05$ \\
        $i = 50$ & $0.76\pm 0.5$ & $1.48 \pm 0.01$ & $1.55 \pm 0.01$ \\
        $i = 100$ & $1.43\pm 0.05$ & $1.53 \pm 0.01$ & $1.49 \pm 0.01$ \\
        \midrule
        \multicolumn{4}{c}{Spot the Cow} \\
        \midrule
        $i = 10$ & $0.54\pm 0.1$ & $0.87 \pm 0.07$ & $1.02 \pm 0.06$ \\
        $i = 50$ & $0.76\pm 0.04$ & $0.95 \pm 0.16$ & $1.08 \pm 0.05$ \\
        $i = 100$ & $0.83\pm 0.1$ & $1.08 \pm 0.05$ & $1.29 \pm 0.05$ \\
        \bottomrule
    \end{tabular}
\end{table}

%% file: Sections/experiments.tex
\section{Experiments}
\label{sec:experiments}

\begin{figure}[bt]
    \centering
    \input{Figures/all-demos.tex}
    \caption{Four experiment of drawing robot; for each experiment we report the demonstrated path (obtained by drawing on a tablet), the trajectory generalised on the target mesh, and the corresponding execution with the robot.}
    \label{fig:dmp-generalisation}
\end{figure}
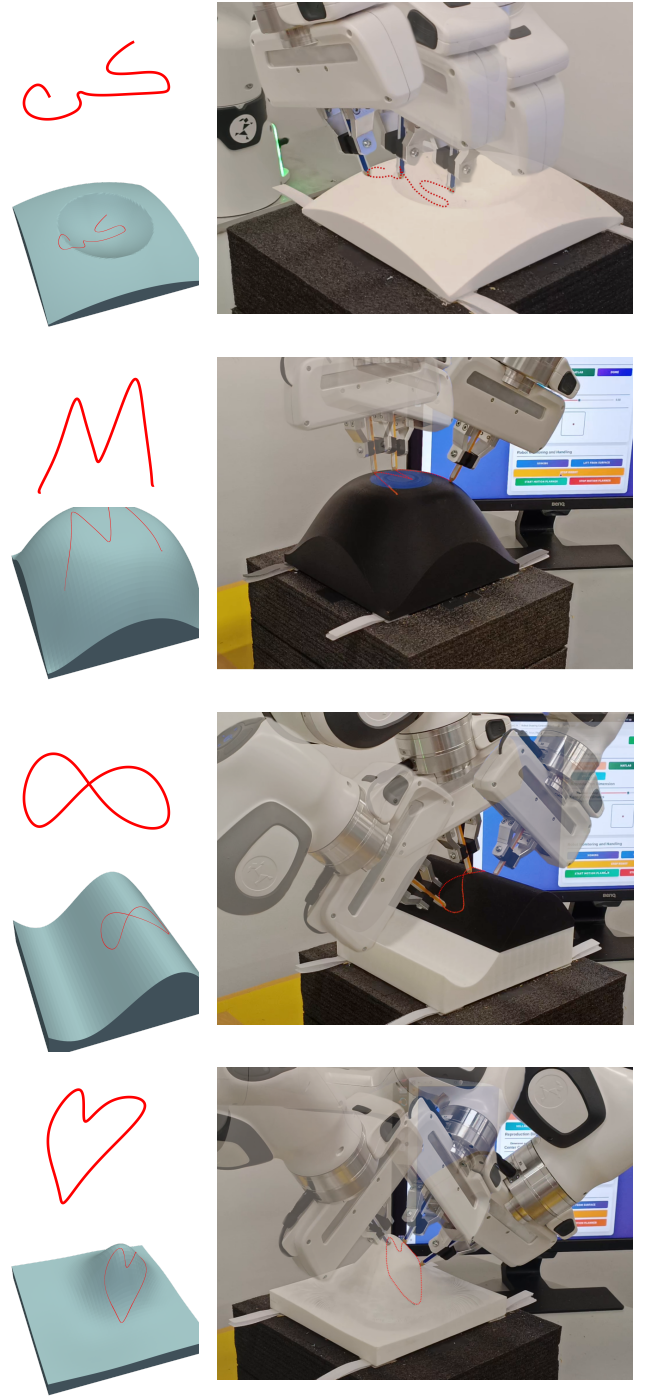

\subsection{Handwritten symbols learning and  generalisation}
\label{sec:handwritten}
The key feature of DMPs is their high generalisation capability, which enables the execution of motion policies to new, unseen initial and final condition while preserving the topology of the learned path by exploiting spatial and temporal invariance properties.
As highlighted in~\eqref{eq:newmeshdmp} and~\eqref{eq:rhythdmp}, exploiting the definition of surface operators on discrete manifolds, it is possible to learn motion policies that can be transferred to different meshes. 
We show this with an experiment in which user-acquired demonstration are generalised to different surfaces to enable a robotic manipulator drawing on them.

For this experiment, we created 4 surfaces generated as the plot of continuous functions of type $H: \R^2\rightarrow \R$; upon fixing a triangulated grid on the $20\times 20\,$cm squared domain, we construct the discrete approximation of the surface $(x, y) \mapsto \big(x, y, H(x,y)\big)$, and we 3D printed these samples.
To recreate a realistic scenario of a smooth surface, the 3D printed components is generated with a high polygonal density, with the triangulated grid having $500\times500$ points; during experiments, instead, we relied on a coarser mesh defined on grids ranging from $40\times 40$ to $80\times 80$ points.
To enable robot motion on the surface, the mesh samples have been localised relative to the base of the Franka Research 3 manipulator~\cite{haddadin2022franka} that we used as robotic platform.
To compensate localisation inaccuracies, we employed an hybrid force-impedance control; particularly, the robot was prompted to track a $5\,$N force along the pencil direction, while other axis were set on impedance control with linear stiffness of $4\,000\,$N/m, rotational stiffness $100\,$Nm/rad, and critical damping.

To ease robot teaching, we designed a simple and intuitive web-based Graphical User Interface (GUI) in which non-expert users can sketch shapes on a tablet, and then decide which type of DMP to use through buttons.
The system immediately learns the flat demonstration and generalises it to all target meshes, creating $40\,000$ trajectory samples typically in less then $5\,$s; 
the GUI also enables updating DMPs parameters---such as starting and final configuration, or motion centre---and control the robot behaviour, like starting or stopping the execution of the motion on a specific surface.
Since the proposed DMPs on surfaces provide only position information for the robots end effector, we select the orientation of the pencil to be normal to the surface by exploiting the normal map given by the mesh itself.
This choice leads to non-differentiable orientation trajectories when transitioning between faces; for this reason, we firstly precompute the whole trajectory, and then use a symmetric sliding-window filter to smoothen the orientation component.

Figure~\ref{fig:dmp-generalisation} effectively shows how the proposed operators on meshes, paired with DMPs~\eqref{eq:newmeshdmp} and~\eqref{eq:rhythdmp}, can generalise the motion from the flat plane to the different types of surfaces.
Since the operators require no smoothness condition of the mesh $\mesh$, our method works also when the underlying function $H$ is non-differentiable in some point locus, as in the case depicted in the last column of that figure. We refer to the accompanying video for more experiments.

\subsection{Surface polishing}

\begin{figure}[bt]
    \centering
    \input{Figures/unitn/polishing.tex}
    \caption{RGB image acquired from the end-effector mounted camera with polynomial path automatically computed by computer-vision algorithms (a). 
    Point cloud with polishing-like trajectory generated on the surface (b). 
Surface before (c) and after (d) the polishing operation, and snapshot of the robot (e) executing the polishing behaviour.
    }
    \label{fig:polishing}
\end{figure}
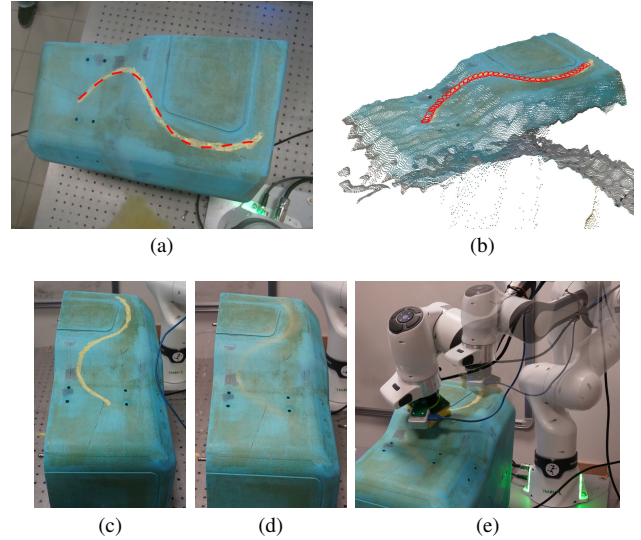

In this experiment, we show how our methodology can be easily integrated in a larger framework for fully autonomous surface polishing, where the surface is reconstructed from sensory data. We consider the setup in Fig.~\ref{fig:polishing}, where the robot is equipped with an Intel Realsense D405 RGB-D camera. The depth camera is used to generate a point cloud of the observed scene, which enables online surface reconstruction and, through simple computer vision algorithms, the generation of polishing motions on the target surface using our rhythmic DMP.

As target workpiece, we use a blue-coloured component with non-flat
surface, and draw a line on it with a orange chalk; by using range
thresholding on the HSL-channels of the colour image, we masked the
drawn line and proceeded fitting a polynomial on it, as shown in
Fig.~\ref{fig:polishing}(a).  To improve mesh reconstruction
reliability, we use a waterfront expanding algorithm to segment the
full point cloud in the region of interested given by the polynomial
path.  To further reduce computational burden, we used Voxel
downsampling with cell size of $5\,$mm on the segmented point cloud,
and then proceeded in using the Poisson surface reconstruction
algorithm~\cite{Kazhdan2006} to generate a watertight, manifold mesh
that we can use for planning.  In our experiments, the proposed
pipeline takes from 1 to 10 seconds, based on point-cloud quality; for
the experiment shown in Fig.~\ref{fig:polishing}, the generated mesh
had $1\,344$ vertices and $2\,617$ faces.

To achieve the polishing behaviour over the surface, we first trained
a rhythmic DMP~\eqref{eq:rhythdmp} on a circular motion on the flat
surface.  We then generalised the policy to the point
cloud-reconstructed mesh, and achieve a polishing-like behaviour by
slowly shifting the DMP centre $g$ along the properly projected fitted
polynomial.  Fixed the velocity of the shifting centre along the path
to $5\,$mm/s and a time-per-revolution of $2\,$s, the algorithm
generates offline the full position trajectory at $100\,$Hz typically
in less then five seconds.  The resulting motion is overlaid on the
acquired point cloud in Fig.~\ref{fig:polishing}(b).

For robotic execution, we equipped the end-effector with a soft sponge, and leverage the hybrid force-impedance controller to exert a constant force on the surface.
The generated trajectory is upsampled to $1\,$kHz and, as in Sec.~\ref{sec:handwritten}, we assign the orientation to be normal to the surface;  to ensure smoothness of the orientation signal, we used a symmetric sliding window of $1\,$s.
Figure~\ref{fig:polishing}(c)--(e) reports the surface before and after the robot polishing, showing how the proposed methodology can be effectively integrated in real tasks where the object surface is not known upfront.

%% file: Figures/all-demos.tex
\begin{tikzpicture}[every node/.style={font=\footnotesize}]

    \def\offy{47mm}

    \newcommand\drawexperiment[2]{
        \node[align=center, anchor=south east, yshift=-2mm] { \includegraphics{Figures/#2/demonstration.pdf}};
        \node[align=center, anchor=north east, yshift= 2mm] { \includegraphics[width=25mm]{Figures/#2/#1.png}};
        \node[align=center, anchor=west]                    { \includegraphics[width=55mm]{Figures/robot-paths/#1-#2.png}};
    }

    \begin{scope}
        \drawexperiment{volcano}{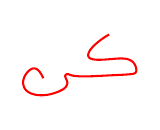}
    \end{scope}
    \begin{scope}[yshift=-\offy]
        \drawexperiment{dome}{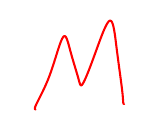}
    \end{scope}
    \begin{scope}[yshift=-2*\offy]
        \drawexperiment{wave}{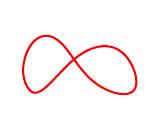}
    \end{scope}
    \begin{scope}[yshift=-3*\offy]
        \drawexperiment{matlab}{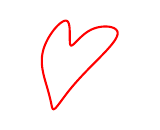}
    \end{scope}
\end{tikzpicture}

%% file: Figures/unitn/polishing.tex
\begin{tikzpicture}[every node/.style={font=\footnotesize}]

    \begin{scope}
        \node[anchor=east, align=center] { \includegraphics[height=3cm]{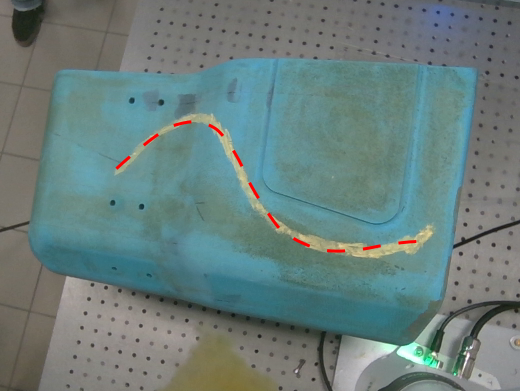} \\ (a) };
        \node[anchor=west, align=center] { \includegraphics[height=3cm]{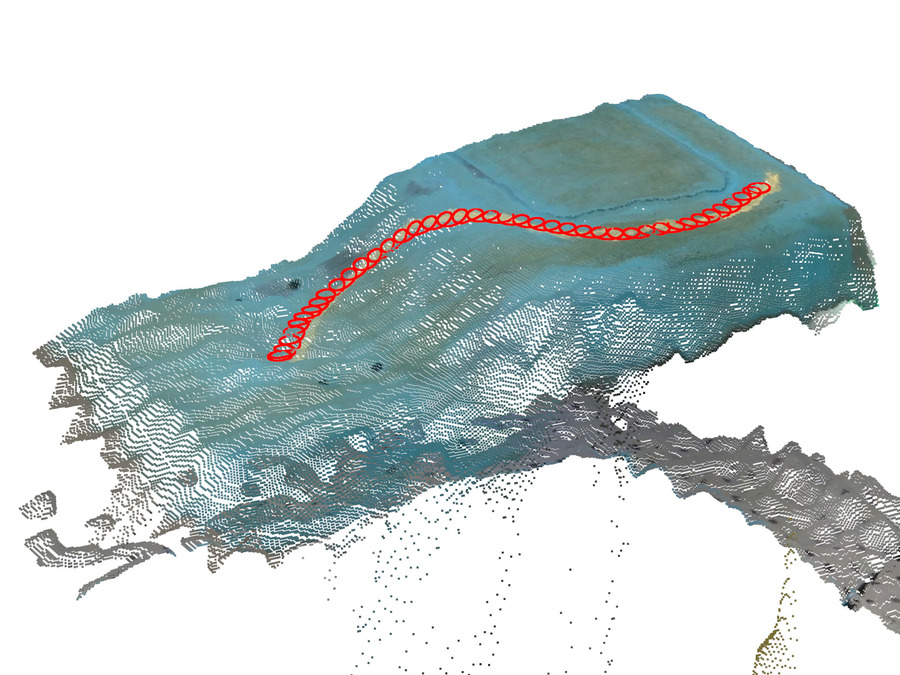}  \\ (b) };
    \end{scope}
    
    \begin{scope}[yshift=-37mm, xshift=-28mm]
        \node[align=center]                            (PRE)  { \includegraphics[height=3cm]{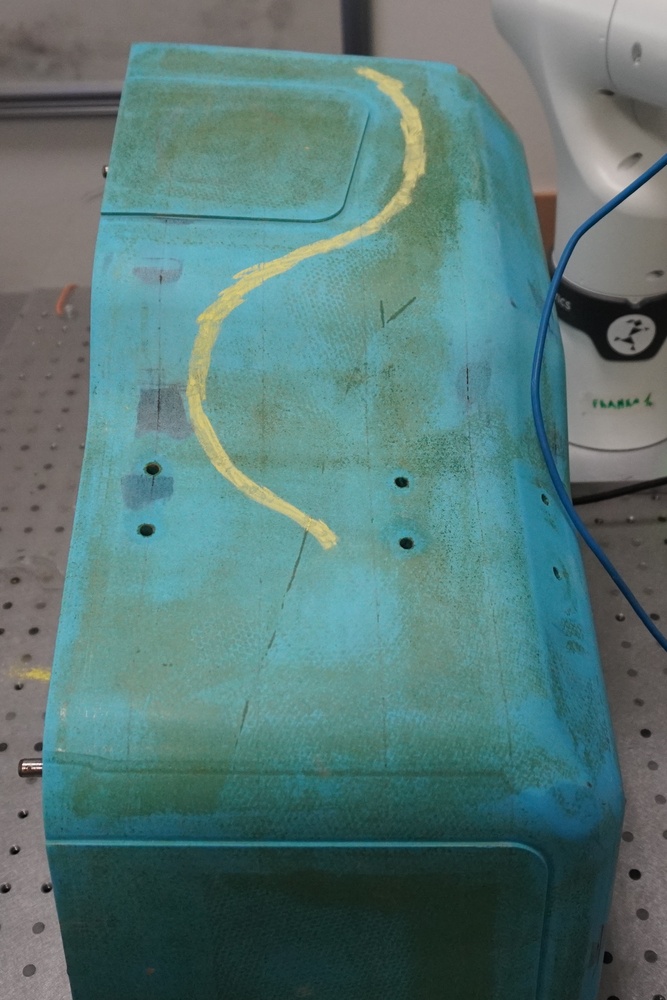}   \\ (c) };
        \node[right of=PRE, anchor=west, align=center] (POST) { \includegraphics[height=3cm]{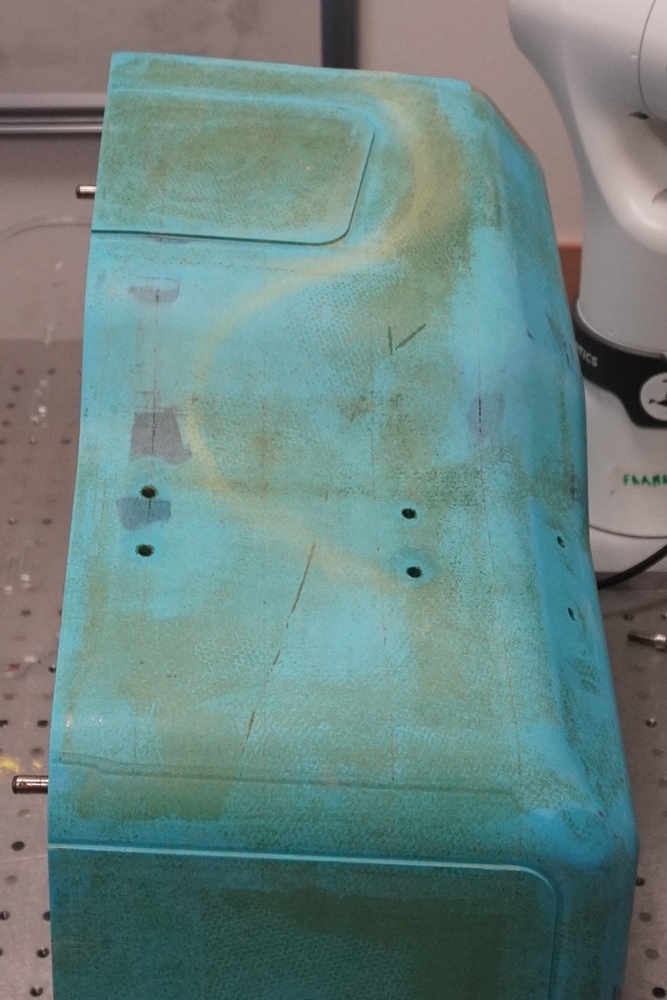}  \\ (d) };
        \node[right of=POST, anchor=west, align=center]       { \includegraphics[height=3cm]{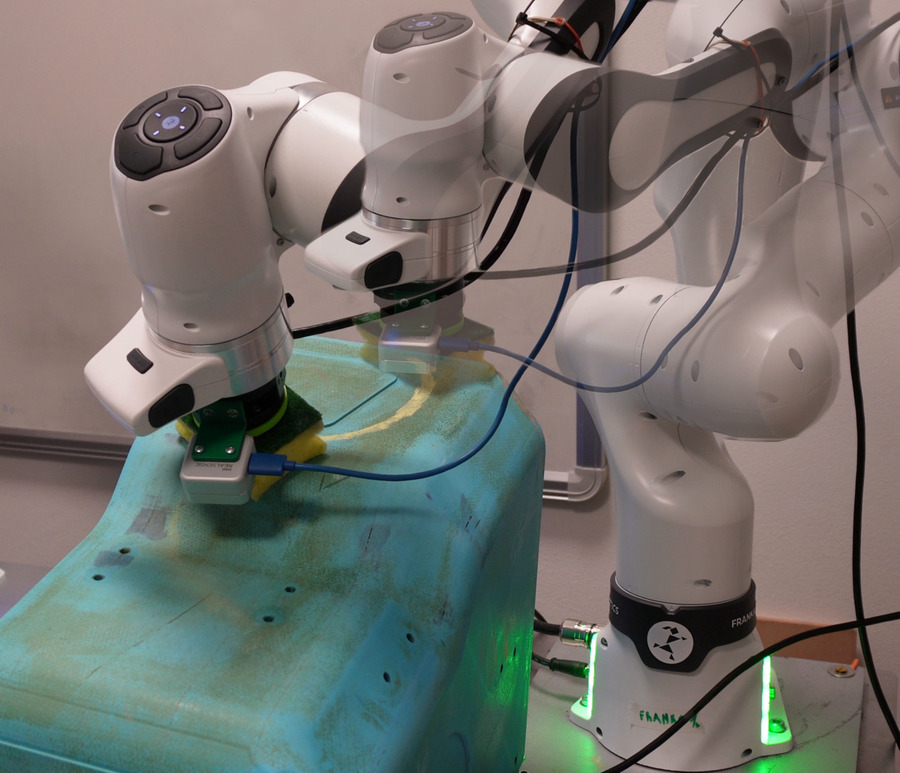} \\ (e) };
    \end{scope}
    
\end{tikzpicture}

%% file: Sections/conclusions.tex
\section{Discussion and Conclusion}
\label{sec:conclusion}
In this work, we developed a suite of geometrically consistent
operators---logarithmic and exponential map, parallel transport, and
ambient space projections---that extend Riemannian differential
geometry to the realm of discrete surfaces modelled by polyhedral
meshes. The core of our approach is to work directly on the intrinsic
geometry of the surfaces, rather than considering the surfaces as mere
constraints to be enforced \textit{a posteriori}. We successfully
applied these operators to a diverse set of geometry-aware learning
methods relevant to the robotics community, namely DMPs, GPs, and RFM.

Concerning DMPs, by encoding the forcing term within a fixed tangent
cone and reconstructing it through parallel transport, we eliminate
the limit-cycle instability of prior approaches, constructing motion
policies that generalise better across multiple surfaces.  We
validated our approach in simulation---comparing against the MeshDMP
framework~\cite{DalleVedove25}---and on a real end-to-end LfD
pipeline, with a Franka Research 3 manipulator reproducing
tablet-drawn trajectories. Both the simulation and the real-world
experiment demonstrate the numerical efficiency of the approach.  As
regards GPs, replacing spectral Laplace-Beltrami
measures~\cite{Borovitskiy2020} with a geodesic-based kernel removes
all smoothness assumptions for the underlying surface, enabling
regression at arbitrary locations on the surface without
interpolation. This technique is particularly effective on coarse
meshes.  Finally, for RFMs, the integration of mesh-specific operators
improves generative quality compared to spectral baselines, whilst
reducing training time by approximately half.

Taken together, these results suggest a remarkable advance towards
bridging geometry-aware learning theory and the mesh representations
employed in robotics.  The cornerstone of our construction is a
principled application of discrete differential geometry.


\paragraph*{Limitations} 
Although we believe that the advances brought about by our approach
are substantial, some of the present limitations should not be
overlooked, as they also offer interesting opportunities for future
investigations.  First, the geodesic pre-computation remains a
bottleneck. Since it relies on algorithms with complexity
$\mathcal{O}(N^2 \log N)$, where $N$ is the number of faces of the
mesh, this computation becomes a significant barrier when the surface
exceeds $10\,000$ faces.

Second, the geodesic-based kernel cannot be guaranteed to be
positive-definite when the lengthscale exceeds some critical value
$\kappa_{th}$. This unavoidable problem comes with the risk of
generating inconsistent GPs when the kernel is evaluated beyond the
critical value.  While Algorithm~\ref{alg:kernel-bisection} offers a
practical safeguard, a better characterisation of the threshold
$\kappa_{th}$ is required---particularly by relating this coefficient
to surface curvature and topology, as well as the number of sampled
points.

\paragraph*{Future directions}
Whilst this paper presents a strong mathematical foundation and three important 
application paradigms, we believe that much remains to be done to extend the 
framework to a broader set of applications.

One promising research avenue is to tackle the challenges posed by the extension of the formulation to 
learn force trajectories on surfaces and, at the same time, to transfer them  across surfaces with different topologies. 
Another open research direction is to extend the framework to time-varying surfaces,  e.g., to enable motion planning for surgical applications on human bodies whose pose  is estimated online via parameterised meshes that evolve over time  with the patient's movements.

%% file: references.bib
@Book{doCarmoRiemann,
  author    = {do Carmo, Manfredo Perdigão},
  publisher = {Birkhäuser},
  title     = {Riemannian geometry},
  year      = {2013},
  address   = {Boston},
  edition   = {14., corrected print.},
  isbn      = {3764334908},
  note      = {Aus dem Portug. übers. - Literaturverz. S. [292] - 296},
  series    = {Mathematics: theory \& applications},
  pagetotal = {300},
  ppn_gvk   = {1615357203},
}

@article{haddadin2022franka,
  title={The franka emika robot: A reference platform for robotics research and education},
  author={Haddadin, Sami and Parusel, Sven and Johannsmeier, Lars and Golz, Saskia and Gabl, Simon and Walch, Florian and Sabaghian, Mohamadreza and J{\"a}hne, Christoph and Hausperger, Lukas and Haddadin, Simon},
  journal={IEEE Robotics \& Automation Magazine},
  volume={29},
  number={2},
  pages={46--64},
  year={2022}
}

@Book{doCarmoSurface,
  author    = {do Carmo, Manfredo Perdigão},
  publisher = {Dover Publications,Inc.},
  title     = {Differential geometry of curves and surfaces},
  year      = {2016},
  address   = {Mineola, New York},
  edition   = {Revised \& updated second edition},
  isbn      = {9780486817972},
  series    = {Mathematics},
  pagetotal = {1510},
  ppn_gvk   = {877923817},
}

@Article{Mitchell1987,
  author    = {Mitchell, Joseph S. B. and Mount, David M. and Papadimitriou, Christos H.},
  journal   = {SIAM Journal on Computing},
  title     = {The Discrete Geodesic Problem},
  year      = {1987},
  issn      = {1095-7111},
  month     = aug,
  number    = {4},
  pages     = {647--668},
  volume    = {16},
  doi       = {10.1137/0216045},
  publisher = {Society for Industrial & Applied Mathematics (SIAM)},
}

@InProceedings{Chen1990,
  author     = {Chen, Jindong and Han, Yijie},
  booktitle  = {Proceedings of the sixth annual symposium on Computational geometry - SCG ’90},
  title      = {Shortest paths on a polyhedron},
  year       = {1990},
  pages      = {360--369},
  publisher  = {ACM Press},
  series     = {SCG ’90},
  collection = {SCG ’90},
  doi        = {10.1145/98524.98601},
}

@Article{Xin2009,
  author    = {Xin, Shi-Qing and Wang, Guo-Jin},
  journal   = {ACM Transactions on Graphics},
  title     = {Improving Chen and Han’s algorithm on the discrete geodesic problem},
  year      = {2009},
  issn      = {1557-7368},
  month     = aug,
  number    = {4},
  pages     = {1--8},
  volume    = {28},
  doi       = {10.1145/1559755.1559761},
  publisher = {Association for Computing Machinery (ACM)},
}

@ARTICLE{Hassler1944,
  title     = "The self-intersections of a smooth n-manifold in 2n-space",
  author    = "Hassler, Whitney",
  journal   = "Ann. Math.",
  publisher = "JSTOR",
  volume    =  45,
  number    =  2,
  pages     = "220",
  month     =  apr,
  year      =  1944
}

@InProceedings{Polthier2006,
  author     = {Polthier, Konrad and Schmies, Markus},
  booktitle  = {ACM SIGGRAPH 2006 Courses on - SIGGRAPH ’06},
  title      = {Straightest geodesics on polyhedral surfaces},
  year       = {2006},
  pages      = {30},
  publisher  = {ACM Press},
  series     = {SIGGRAPH ’06},
  collection = {SIGGRAPH ’06},
  doi        = {10.1145/1185657.1185664},
}

@ARTICLE{Ijspeert13,
    author={Ijspeert, Auke Jan and Nakanishi, Jun and Hoffmann, Heiko and Pastor, Peter and Schaal, Stefan},
    journal={Neural Computation}, 
    title={Dynamical Movement Primitives: Learning Attractor Models for Motor Behaviors}, 
    year={2013},
    volume={25},
    number={2},
    pages={328-373},
    doi={10.1162/NECO_a_00393}
}

@article{saveriano2023dmp,
    title = "Dynamic movement primitives in robotics: A tutorial survey",
    author = "Matteo Saveriano and Abu-Dakka, {Fares J.} and Alja{\v z} Kramberger and Luka Peternel",
    year = "2023",
    volume = "42",
    pages = "1133--1184",
    journal = "The International Journal of Robotics Research",
    issn = "0278-3649",
    publisher = "SAGE Publishing",
    number = "13",
}

@INPROCEEDINGS{Ude14,
    author={Ude, Aleš and Nemec, Bojan and Petrić, Tadej and Morimoto, Jun},
    booktitle={2014 IEEE International Conference on Robotics and Automation (ICRA)}, 
    title={Orientation in Cartesian space dynamic movement primitives}, 
    year={2014},
    volume={},
    number={},
    pages={2997-3004},
    doi={10.1109/ICRA.2014.6907291}
}

@InProceedings{Koutras20a,
    title = 	 {A correct formulation for the Orientation Dynamic Movement Primitives for robot control in the Cartesian space},
    author =       {Koutras, Leonidas and Doulgeri, Zoe},
    booktitle = 	 {Proceedings of the Conference on Robot Learning},
    pages = 	 {293--302},
    year = 	 {2020},
    editor = 	 {Kaelbling, Leslie Pack and Kragic, Danica and Sugiura, Komei},
    volume = 	 {100},
    series = 	 {Proceedings of Machine Learning Research},
    month = 	 {30 Oct--01 Nov},
    publisher =    {PMLR},
    url = 	 {https://proceedings.mlr.press/v100/koutras20a.html},
}

@INPROCEEDINGS{AbuDakka20,
    author={Abu-Dakka, Fares J. and Kyrki, Ville},
    booktitle={2020 IEEE International Conference on Robotics and Automation (ICRA)}, 
    title={Geometry-aware Dynamic Movement Primitives}, 
    year={2020},
    volume={},
    number={},
    pages={4421-4426},
    doi={10.1109/ICRA40945.2020.9196952}
}

@article{AbuDakka24,
    title = {A unified formulation of geometry-aware discrete dynamic movement primitives},
    journal = {Neurocomputing},
    volume = {598},
    pages = {128056},
    year = {2024},
    issn = {0925-2312},
    doi = {https://doi.org/10.1016/j.neucom.2024.128056},
    author = {Fares J. Abu-Dakka and Matteo Saveriano and Ville Kyrki},
}

@INPROCEEDINGS{DalleVedove25,
    author={Dalle Vedove, Matteo and Abu-Dakka, Fares J. and Palopoli, Luigi and Fontanelli, Daniele and Saveriano, Matteo},
    booktitle={2025 IEEE International Conference on Robotics and Automation (ICRA)}, 
    title={MeshDMP: Motion Planning on Discrete Manifolds Using Dynamic Movement Primitives}, 
    year={2025},
    volume={},
    number={},
    pages={895-901},
    doi={10.1109/ICRA55743.2025.11128556}
}

@article{Xue2023,
    author = {Xue, Xianfa and Dong, Jiale and Lu, Zhenyu and Wang, Ning},
    title = {A robotic learning and generalization framework for curved surface based on modified DMP},
    journal = {Robotics and Autonomous Systems},
    volume = {160},
    pages = {104323},
    year = {2023}
}

@article{Berger2017,
    title = {A Survey of Surface Reconstruction from Point Clouds},
    volume = {36},
    number = {1},
    journal = {Computer Graphics Forum},
    publisher = {Wiley},
    author = {Berger,  Matthew and Tagliasacchi,  Andrea and Seversky,  Lee M. and Alliez,  Pierre and Guennebaud,  Gaël and Levine,  Joshua A. and Sharf,  Andrei and Silva,  Claudio T.},
    year = {2017},
    pages = {301–329}
}

@Book{Rasmussen2005,
    author    = {Rasmussen, Carl Edward and Williams, Christopher K. I.},
    publisher = {The MIT Press},
    title     = {Gaussian Processes for Machine Learning},
    year      = {2005},
    isbn      = {9780262256834},
    month     = nov,
    doi       = {10.7551/mitpress/3206.001.0001},
}

@article{Calinon20RAM,
	author="Calinon, S.",
	title="Gaussians on {R}iemannian Manifolds: Applications for Robot Learning and Adaptive Control",
	journal="{IEEE} Robotics and Automation Magazine ({RAM})",
	year="2020",
	month="June",
	volume="27",
	number="2",
	pages="33--45",
	doi="10.1109/MRA.2020.2980548"
}

@Article{Kniely2014,
    author    = {Kniely, Michael and Ring, Wolfgang},
    journal   = {Inverse Problems in Science and Engineering},
    title     = {Riemannian methods for optimization in a shape space of triangular meshes},
    year      = {2014},
    issn      = {1741-5985},
    month     = nov,
    number    = {6},
    pages     = {1011--1039},
    volume    = {23},
    doi       = {10.1080/17415977.2014.980243},
    publisher = {Informa UK Limited},
}

@inproceedings{Borovitskiy2020,
    author = {Borovitskiy, Viacheslav and Terenin, Alexander and Mostowsky, Peter and Deisenroth, Marc Peter},
    title = {Mat\'{e}rn Gaussian processes on riemannian manifolds},
    year = {2020},
    isbn = {9781713829546},
    publisher = {Curran Associates Inc.},
    address = {Red Hook, NY, USA},
    booktitle = {Proceedings of the 34th International Conference on Neural Information Processing Systems},
    articleno = {1042},
    numpages = {12},
    location = {Vancouver, BC, Canada},
    series = {NIPS '20}
}

@inproceedings{DeFloriani2005,
    author = {De Floriani, Leila and Hui, Annie},
    title = {Data structures for simplicial complexes: an analysis and a comparison},
    year = {2005},
    isbn = {390567324X},
    publisher = {Eurographics Association},
    address = {Goslar, DEU},
    booktitle = {Proceedings of the Third Eurographics Symposium on Geometry Processing},
    pages = {119–es},
    location = {Vienna, Austria},
    series = {SGP '05}
}

@InProceedings{Feragen2015,
    author    = {Feragen, Aasa and Lauze, Francois and Hauberg, Soren},
    booktitle = {IEEE Conference on Computer Vision and Pattern Recognition},
    title     = {Geodesic exponential kernels: When curvature and linearity conflict},
    year      = {2015},
    month     = jun,
    pages     = {3032--3042},
    publisher = {IEEE},
    doi       = {10.1109/cvpr.2015.7298922},
}

@article{mostowsky2024,
    title = {The GeometricKernels Package: Heat and Matérn Kernels for Geometric Learning on Manifolds, Meshes, and Graphs},
    author = {Peter Mostowsky and Vincent Dutordoir and Iskander Azangulov and Noémie Jaquier and Michael John Hutchinson and Aditya Ravuri and Leonel Rozo and Alexander Terenin and Viacheslav Borovitskiy},
    year = {2024},
    journal = {arXiv:2407.08086},
}

@inproceedings{StanfordDragon,
  title      = "A volumetric method for building complex models from range
                images",
  booktitle  = "Proceedings of the 23rd annual conference on Computer graphics
                and interactive techniques",
  author     = "Curless, Brian and Levoy, Marc",
  publisher  = "ACM",
  year       =  "1996",
  conference = " International Conference on Computer Graphics
                and Interactive Techniques"
}

@ARTICLE{Sentis2005,
    title     = "Synthesis of whole-body behaviors through hierarchical control of behavioral primitives",
    author    = "Sentis, Luis and Khatib, Oussama",
    journal   = "International Journal of Humanoid Robotics",
    publisher = "World Scientific Pub Co Pte Lt",
    volume    =  "2",
    number    =  "4",
    pages     = "505--518",
    year      =  "2005",
    language  = "en"
}

@ARTICLE{Jaquier2021,
  title     = "Geometry-aware manipulability learning, tracking, and transfer",
  author    = "Jaquier, No{\'e}mie and Rozo, Leonel and Caldwell, Darwin G and
               Calinon, Sylvain",
  journal   = "International Journal of Robotics Research (IJRR)",
  publisher = "SAGE Publications",
  volume    =  40,
  number    = "2-3",
  pages     = "624--650",
  month     =  feb,
  year      =  2021,
  language  = "en"
}

@INPROCEEDINGS{Chen2023,
    title        = "Flow Matching on general geometries",
    author       = "Chen, Ricky T Q and Lipman, Yaron",
    booktitle    = "12th International Conference on Learning Representations (ICLR)", 
    year         =  2024,
    eprint       = "2302.03660"
}

@TECHREPORT{Hubeli2000,
    title     = "A survey of surface representations for geometric modeling",
    author    = "Hubeli, Andreas and Gross, Markus",
    publisher = "ETH Zurich, Department of Computer Science",
    institution  = "ETH Zurich, Department of Computer Science",
    year      =  "2000"
}

@BOOK{Bobenko2023,
  title     = "Discrete differential geometry: Integrable structure",
  author    = "Bobenko, Alexander I and Suris, Yuri B",
  publisher = "American Mathematical Society",
  year      =  "2023",
  address   = "Providence, RI",
  language  = "en"
}

@article{billard2016learning,
  title={Learning from humans},
  author={Billard, Aude G and Calinon, Sylvain and Dillmann, R{\"u}diger},
  journal={Springer handbook of robotics},
  pages={1995--2014},
  year={2016},
  publisher={Springer}
}

@article{huang2019kernelized,
    title={Kernelized movement primitives},
    author={Huang, Yanlong and Rozo, Leonel and Silv{\'e}rio, Joao and Caldwell, Darwin G},
    journal={The International Journal of Robotics Research},
    volume={38},
    number={7},
    pages={833--852},
    year={2019},
    publisher={SAGE Publications Sage UK: London, England}
}

@article{calinon2016tutorial,
    title={A tutorial on task-parameterized movement learning and retrieval},
    author={Calinon, Sylvain},
    journal={Intelligent service robotics},
    volume={9},
    pages={1--29},
    year={2016}
}

@article{paraschos2013,
    title={Probabilistic movement primitives},
    author={Paraschos, Alexandros and Daniel, Christian and Peters, Jan R and Neumann, Gerhard},
    journal={Advances in neural information processing systems},
    volume={26},
    year={2013}
}

@ARTICLE{Kim2025,
    author={Kim, Jonghyeok and Sung, Minchang and Choi, Youngjin and Park, Jonghoon and Chung, Wan Kyun},
    journal={IEEE Transactions on Robotics}, 
    title={Impedance Control Design Framework Using Commutative Map Between $SE(3)$ and $\mathfrak {se}(3)$}, 
    year={2025},
    volume={41},
    number={},
    pages={6193-6212},
    doi={10.1109/TRO.2025.3619066}
}

@ARTICLE{AbuDakka2020-vicl,
  title     = "Variable impedance control and learning-A review",
  author    = "Abu-Dakka, Fares J and Saveriano, Matteo",
  journal   = "Front. Robot. AI",
  publisher = "Frontiers Media SA",
  volume    =  7,
  pages     = "590681",
  month     =  dec,
  year      =  2020,
  copyright = "https://creativecommons.org/licenses/by/4.0/",
  language  = "en"
}

@misc{Ratliff2018,
    Author = {Ratliff, Nathan D. and Issac, Jan and Kappler, Daniel and Birchfield, Stan and Fox, Dieter},
    Title = {Riemannian Motion Policies},
    Year = {2018},
    Eprint = {arXiv:1801.02854},
}

@INPROCEEDINGS{Crane2013-hp,
  title           = "Digital geometry processing with discrete exterior
                     calculus",
  booktitle       = "{ACM} {SIGGRAPH} 2013 Courses",
  author          = "Crane, Keenan and de Goes, Fernando and Desbrun, Mathieu
                     and Schr{\"o}der, Peter",
  publisher       = "ACM",
  month           =  jul,
  year            =  2013,
  address         = "New York, NY, USA",
  conference      = "SIGGRAPH '13: Special Interest Group on Computer Graphics
                     and Interactive Techniques Conference",
  location        = "Anaheim California"
}

@ARTICLE{Vallet2008-al,
  title     = "Spectral geometry processing with Manifold Harmonics",
  author    = "Vallet, B and L{\'e}vy, B",
  journal   = "Comput. Graph. Forum",
  publisher = "Wiley",
  volume    =  27,
  number    =  2,
  pages     = "251--260",
  month     =  apr,
  year      =  2008,
  copyright = "http://onlinelibrary.wiley.com/termsAndConditions\#vor",
  language  = "en"
}

@article{Patane2014,
    title = {Laplacian spectral distances and kernels on 3D shapes},
    journal = {Pattern Recognition Letters},
    volume = {47},
    pages = {102-110},
    year = {2014},
    note = {Advances in Mathematical Morphology},
    issn = {0167-8655},
    doi = {https://doi.org/10.1016/j.patrec.2014.04.003},
    url = {https://www.sciencedirect.com/science/article/pii/S0167865514001214},
    author = {Giuseppe Patané},
}

@ARTICLE{Lang2023,
  title     = "{Galerkin--Chebyshev} approximation of Gaussian random fields on
               compact Riemannian manifolds",
  author    = "Lang, Annika and Pereira, Mike",
  journal   = "BIT",
  publisher = "Springer Science and Business Media LLC",
  volume    =  63,
  number    =  4,
  month     =  dec,
  year      =  2023,
  copyright = "https://creativecommons.org/licenses/by/4.0",
  language  = "en"
}

@InProceedings{Koestler2022,
    author="Koestler, Lukas
    and Grittner, Daniel
    and Moeller, Michael
    and Cremers, Daniel
    and L{\"a}hner, Zorah",
    editor="Avidan, Shai
    and Brostow, Gabriel
    and Ciss{\'e}, Moustapha
    and Farinella, Giovanni Maria
    and Hassner, Tal",
    title="Intrinsic Neural Fields: Learning Functions on Manifolds",
    booktitle="European Conference on Computer Vision",
    year="2022",
    publisher="Springer Nature Switzerland",
    address="Cham",
    pages="622--639",
}

@inproceedings{Lipman2023,
    title={Flow Matching for Generative Modeling},
    author={Yaron Lipman and Ricky T. Q. Chen and Heli Ben-Hamu and Maximilian Nickel and Matthew Le},
    booktitle={The Eleventh International Conference on Learning Representations },
    year={2023},
}

@inproceedings{Liu2023,
    title={Flow straight and fast: Learning to generate and transfer data with rectified flow},
    author={Liu, Xingchao and Gong, Chengyue and Liu, Qiang},
    booktitle={The Eleventh International Conference on Learning Representations },
    year={2023},
}

@inproceedings{Boier2004,
    author = {Boier-Martin, Ioana and Zorin, Denis},
    title = {Differentiable parameterization of Catmull-Clark subdivision surfaces},
    year = {2004},
    isbn = {3905673134},
    publisher = {Association for Computing Machinery},
    url = {https://doi.org/10.1145/1057432.1057453},
    doi = {10.1145/1057432.1057453},
    booktitle = {ACM Symposium on Geometry Processing},
    pages = {155--164}
}

@ARTICLE{Dimas1999-cp,
    title     = "{3D} geometric modelling based on {NURBS}: a review",
    author    = "Dimas, E and Briassoulis, D",
    journal   = "Adv. Eng. Softw.",
    publisher = "Elsevier BV",
    volume    =  30,
    number    = "9-11",
    pages     = "741--751",
    month     =  sep,
    year      =  1999,
    language  = "en"
}

@ARTICLE{Choi2020,
    title     = "Parallelizable global conformal parameterization of
                 simply-connected surfaces via partial welding",
    author    = "Choi, Gary P T and Leung-Liu, Yusan and Gu, Xianfeng and Lui, Lok Ming",
    journal   = "SIAM J. Imaging Sci.",
    publisher = "Society for Industrial \& Applied Mathematics (SIAM)",
    volume    =  13,
    number    =  3,
    pages     = "1049--1083",
    month     =  jan,
    year      =  2020,
    language  = "en"
}

@ARTICLE{Pantic2021,
    author={Pantic, Michael and Ott, Lionel and Cadena, Cesar and Siegwart, Roland and Nieto, Juan},
    journal={IEEE Robotics and Automation Letters}, 
    title={Mesh Manifold Based Riemannian Motion Planning for Omnidirectional Micro Aerial Vehicles}, 
    year={2021},
    volume={6},
    number={3},
    pages={4790-4797},
    doi={10.1109/LRA.2021.3061869}
}

@Book{Clarke2013,
    author    = {Clarke, Francis},
    publisher = {Springer London},
    title     = {Functional Analysis, Calculus of Variations and Optimal Control},
    year      = {2013},
    isbn      = {9781447148203},
    doi       = {10.1007/978-1-4471-4820-3},
    issn      = {2197-5612},
    journal   = {Graduate Texts in Mathematics},
}

@BOOK{Ericson2004,
    title     = "{Real-Time} Collision Detection",
    author    = "Ericson, Christer",
    publisher = "CRC Press",
    series    = "Morgan Kaufmann series in interactive 3D technology",
    month     =  dec,
    year      =  2004,
    address   = "London, England",
    language  = "en"
}

@inproceedings{Chen2018,
    author = {Chen, Ricky T. Q. and Rubanova, Yulia and Bettencourt, Jesse and Duvenaud, David},
    title = {Neural ordinary differential equations},
    year = {2018},
    publisher = {Curran Associates Inc.},
    address = {Red Hook, NY, USA},
    booktitle = {Proceedings of the 32nd International Conference on Neural Information Processing Systems},
    pages = {6572–6583},
    numpages = {12},
    location = {Montr\'{e}al, Canada},
    series = {NIPS'18}
}

@article{Crane2013,
  title={Robust fairing via conformal curvature flow},
  author={Crane, Keenan and Pinkall, Ulrich and Schr{\"o}der, Peter},
  journal={ACM Transactions on Graphics (TOG)},
  volume={32},
  number={4},
  pages={1--10},
  year={2013},
  publisher={ACM New York, NY, USA}
}

@article{Whittle63,
  title={Stochastic processes in serveral dimensions},
  author={Whittle, Peter},
  journal={Bulletin of the International Statistical Institute},
  volume={40},
  pages={974--994},
  year={1963},
}

@inproceedings{Ramachandran18,
  author       = {Prajit Ramachandran and
                  Barret Zoph and
                  Quoc V. Le},
  title        = {Searching for Activation Functions},
  booktitle    = {International Conference on Learning Representations},
  year         = {2018}
}

@book{cgal-library
, title        = "{CGAL} User and Reference Manual"
, author      = "{The CGAL Project}"
, publisher     = "{CGAL Editorial Board}"
, edition     = "{6.1.1}"
, year         = 2026
, url =    "https://doc.cgal.org/6.1.1/Manual/packages.html"
}

@incollection{cgal-shortest-path,
    author = {Stephen Kiazyk and S{\'e}bastien Loriot and {\'E}ric Colin de Verdi{\`e}re},
    title = {Triangulated Surface Mesh Shortest Paths},
    publisher = {{CGAL Editorial Board}},
    edition = {{6.1.1}},
    booktitle = {{CGAL} User and Reference Manual},
    url = {https://doc.cgal.org/6.1.1/Manual/packages.html#PkgSurfaceMeshShortestPath},
    year = 2026
}

@inproceedings{Kazhdan2006,
    author = {Kazhdan, Michael and Bolitho, Matthew and Hoppe, Hugues},
    title = {Poisson surface reconstruction},
    year = {2006},
    isbn = {3905673363},
    publisher = {Eurographics Association},
    address = {Goslar, DEU},
    booktitle = {Proceedings of the Fourth Eurographics Symposium on Geometry Processing},
    pages = {61–70},
    numpages = {10},
    location = {Cagliari, Sardinia, Italy},
    series = {SGP '06}
}
